\PassOptionsToPackage{table}{xcolor}
\documentclass{preprint}
\usepackage[T1]{fontenc}
\usepackage{amsmath,amssymb,graphicx,booktabs,multirow,array}
\usepackage{xcolor}
\usepackage{float}
\usepackage{wrapfig}
\usepackage{needspace}
\usepackage{placeins}
\usepackage{hyperref}
\usepackage{url}
\usepackage{perpage}
\DeclareFontShape{T1}{bytesans}{b}{n}{<-> s * [1] ./bytesans}{}
\MakeSorted{table}
\definecolor{oursblue}{HTML}{0072B2}
\definecolor{ourslight}{HTML}{EDF6FB}

\newcommand{\fdr}{\mathrm{FD}_{r}^{6}}
\newcommand{\norm}[1]{\left\lVert #1\right\rVert}
\newcommand{\LN}{\operatorname{LN}}
\newcommand{\lift}{HiRAE}
\newcommand{\liftfull}{HiRAE-24}
\newsavebox{\tablewidthbox}
\NewDocumentEnvironment{fullwidthtabular}{m m +b}{%
  \begingroup
  \setlength{\tabcolsep}{0pt}%
  \sbox{\tablewidthbox}{\begin{tabular}{#2}#3\end{tabular}}%
  \setlength{\tabcolsep}{\dimexpr(\linewidth-\wd\tablewidthbox)/#1/2\relax}%
  \begin{tabular}{#2}#3\end{tabular}%
  \endgroup
}{}
\title{HiRAE: Hierarchical Representation\\Autoencoding with Residual Budgets}
\renewcommand{\authorlist}{%
  \authorformat[1]{Xuanyu Zhu}, \authorformat[2]{Yan Bai},
  \authorformat[1,\spadesuit]{Yang Shi}, \authorformat[1]{Yihang Lou}\\[2pt]
  \authorformat[1]{Yuanxing Zhang}, \authorformat[1]{Tengfei Liu},
  \authorformat[3]{Jing Jin}, \authorformat[4,\dagger]{Yuan Zhou}%
}
\renewcommand{\affiliationlist}{%
  \affiliationformat[1]{Peking University}\quad
  \affiliationformat[2]{Agibot Research}\\[2pt]
  \affiliationformat[3]{Tsinghua University}\quad
  \affiliationformat[4]{IGDL}%
}
\providecommand{\firstpagefootnotes}{}
\renewcommand{\firstpagefootnotes}{%
  \parbox{0.96\textwidth}{%
    {\color{seedblue}\hrule height 0.2pt\relax}
    \vspace{5pt}
    \raggedright\small
    $^{\spadesuit}$ Project Leader.\qquad $^{\dagger}$ Corresponding Author.
  }%
}
\renewcommand{\titlefigure}{\noindent\begin{minipage}{\linewidth}
\centering
\includegraphics[width=0.80\linewidth]{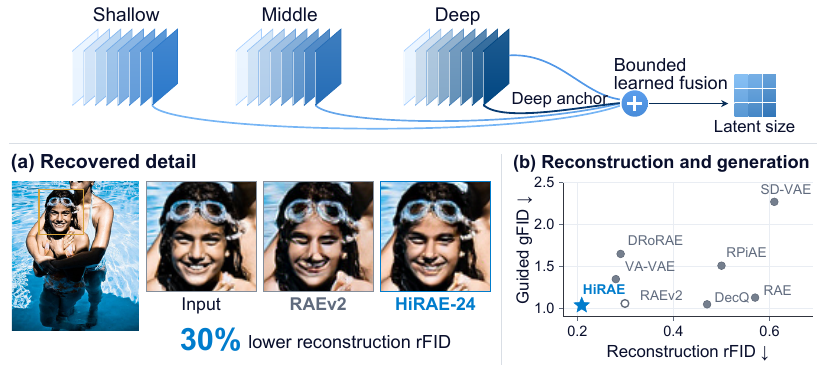}
\captionsetup{hypcap=false}\captionof{figure}{\textbf{Learning full-depth fusion for higher reconstruction fidelity.} Top: HiRAE combines bounded residuals from three encoder-depth groups with the deepest-layer anchor. Left: matched input and reconstruction crops; HiRAE-24 reduces rFID by 30\% relative to RAEv2. Right: reconstruction and guided generation across systems under their respective reported settings.}
\label{fig:overview}
\end{minipage}
\par\vspace{4mm}
}
\abstract{Pretrained visual representations support image generation, but may not fully preserve the fine-grained details needed for faithful reconstruction. Meanwhile, intermediate encoder layers contain complementary visual details, but learning to fuse them for reconstruction can produce a latent distribution that is difficult to model. Existing fusion methods require empirical tuning of layer selection or staged optimization of fusion and decoding, increasing configuration effort or training complexity. We introduce \lift{} (Hierarchical Representation Autoencoder), which learns a hierarchical fusion framework over the full encoder hierarchy to improve reconstruction fidelity while maintaining compatibility with generative modeling. HiRAE groups encoder layers by depth and learns residual corrections to the deepest representation. Group-wise norm caps bound these corrections relative to the deep anchor, with tighter budgets for shallower groups. Our \liftfull{} preserves the latent token count and channel dimension. On ImageNet-256, \liftfull{} reduces reconstruction FID from 0.299 to 0.209 relative to RAEv2 while maintaining competitive guided generation quality.
For text-to-image generation, \liftfull{} improves alignment over RAEv2 on GenEval, DPG-Bench, and GenAI-Bench both before and after supervised fine-tuning. Under the same generator-training and evaluation protocol, post-fine-tuning GenEval increases from 84.86 to 87.70.
}
\hypersetup{pdftitle={HiRAE: Hierarchical Representation Autoencoding with Residual Budgets},pdfauthor={Xuanyu Zhu, Yan Bai, Yang Shi, Yihang Lou, Yuanxing Zhang, Tengfei Liu, Jing Jin, Yuan Zhou}}
\begin{document}
\raggedbottom
\maketitle
\section{Introduction}
Pretrained vision encoders provide semantically organized representations for image generation, but their final outputs can omit details needed for faithful reconstruction~\citep{wang2026decq,singh2026raev2}. Representation Autoencoders (RAE)~\citep{zheng2025rae} pair these frozen encoders with learned decoders and train diffusion models in the resulting latent space. Many previous methods rely solely on the highly abstracted semantic features of the final encoder layer, whereas reconstruction depends more on the detailed features retained in intermediate layers~\citep{singh2026raev2,zhu2026drorae}. Learning to use this information offers a route to higher reconstruction fidelity while retaining the pretrained encoder as the basis for generation.
\par

Recent tokenizers exploit the visual hierarchy to recover details missing from final-layer representations, through fixed aggregation (RAEv2; \citealp{singh2026raev2}), learned full-depth fusion (DRoRAE; \citealp{zhu2026drorae}), or queries over intermediate features (DecQ; \citealp{wang2026decq}). IDEAL~\citep{chen2026ideal} combines selected shallow and deep features before quantization, while LV-RAE~\citep{liu2026lvrae} supplements semantic features with a separate encoder for low-level detail. DecQ shows that using shallower layers or increasing the number of detail queries can improve reconstruction while worsening generation. For learned fusion, this trade-off raises a further concern: reconstruction-driven training can favor shallow-layer detail without accounting for its effect on generation quality. Existing approaches reconcile reconstruction and generation through predefined layer aggregation, additional detail pathways, or staged adaptation of fusion and decoding. Our seven-layer experiments show that learned fusion improves reconstruction while supporting guided generation, but identifying a suitable layer subset requires repeated training and evaluation. We aim to learn a unified representation for reconstruction and generation through joint training of full-hierarchy fusion and the decoder. 
How can we learn full-hierarchy fusion that improves reconstruction while maintaining compatibility with generative modeling?

We introduce \lift{} (Hierarchical Representation Autoencoder), a hierarchical fusion framework that integrates representations across encoder depths into a shared latent space for reconstruction and generation. Its main configuration, \liftfull{}, learns spatially varying contributions from all 24 layers of a frozen DINOv3-L encoder, avoiding manual layer-subset selection. Building on DRoRAE's learned residual fusion~\citep{zhu2026drorae}, HiRAE organizes encoder layers into shallow, middle, and deep groups. Each group learns a residual correction to the deepest representation, with a distinct norm budget that increases with depth. These designs allow us to constrain how fusion modifies the deepest representation and jointly train the fusion module and decoder without a separate fusion-only adaptation phase. The fused representation preserves the original latent token count and channel dimension.
On ImageNet-256~\citep{deng2009imagenet}, \liftfull{} reduces rFID from 0.299 to 0.209, a 30\% reduction relative to RAEv2~\citep{singh2026raev2}. On the matched 5,000-image reconstruction subset, it increases PSNR from 22.667 to 26.377 dB and reduces LPIPS~\citep{zhang2018lpips} from 0.074 to 0.043. After 80 epochs of generator training, guided generation FID decreases from 1.060 to 1.038 (Figure~\ref{fig:overview}). Analysis shows that fusion adds spatial detail while largely preserving class neighborhoods. The learned tokenizer also exhibits lower decoding sensitivity to the tested latent perturbations.

Our contributions are:
\begin{itemize}
\item \textbf{Hierarchical representation autoencoding.} We introduce HiRAE, a hierarchical fusion framework with depth-dependent residual budgets. These budgets control intermediate-layer contributions to enrich the deepest representation with complementary visual detail.
\item \textbf{Higher reconstruction fidelity and improved text-to-image alignment.} \liftfull{} reduces reconstruction FID by 30\% relative to RAEv2 with competitive guided ImageNet generation. Under our shared text-to-image protocol, it improves GenEval, DPG-Bench, and GenAI-Bench scores before and after supervised fine-tuning, with a 2.84-point GenEval gain after fine-tuning.
\item \textbf{HiRAE's latent structure and decoding sensitivity.} Our analysis shows that hierarchical fusion enriches spatial detail while largely preserving class neighborhoods. The learned tokenizer also exhibits lower decoding sensitivity to the tested latent perturbations.
\end{itemize}

\section{Related work}
\paragraph{Visual representations for image generation.}
Latent diffusion models such as LDM and DiT generate images in the compressed spaces of reconstruction-trained autoencoders~\citep{rombach2022ldm,peebles2023dit}.
Representation alignment connects these generative models with pretrained visual encoders at different stages: REPA supervises diffusion features, whereas VA-VAE regularizes the tokenizer latents themselves~\citep{yu2025repa,yao2025vavae}.
Extending this connection to joint optimization, REPA-E uses alignment to support end-to-end tuning of the VAE and diffusion model~\citep{leng2025repae}.
RAE takes a more direct route by pairing a frozen vision encoder with a learned decoder and training diffusion in the encoder's representation space~\citep{zheng2025rae}.
HiRAE extends RAE with a learnable fusion module over the full frozen encoder hierarchy and jointly trains this module with the decoder.

\paragraph{Hierarchical fusion and detail enrichment.}
RAEv2~\citep{singh2026raev2} extends representation autoencoding through fixed aggregation of selected encoder layers, incorporating intermediate-layer detail into the representation used for reconstruction and generation.
The aggregation itself introduces no learned fusion module, making layer selection a key design choice.
Related detail-enrichment designs include shallow-deep fusion before quantization in IDEAL and additional detail representations in DecQ and LV-RAE~\citep{chen2026ideal,wang2026decq,liu2026lvrae}.
For learnable multi-layer fusion, DRoRAE combines layer-wise experts with routing across all encoder layers and trains the fusion module before adapting the decoder~\citep{zhu2026drorae}.
Building on this learned full-depth fusion, HiRAE introduces depth-dependent residual budgets that support joint fusion and decoder training.

\paragraph{Latent structure and generative modeling.}
Adding reconstruction detail also changes the representation that the generator must model, so improvements in reconstruction alone do not establish better generation~\citep{yao2025vavae,wang2026decq}.
FAE and HAE adapt pretrained representations for generation through feature compression and hyperspherical modeling, respectively~\citep{gao2025fae,chang2026hae}.
Complementing these architectural approaches, \citet{zhong2026diffusability} systematically examine how latent properties relate to generation quality across tokenizer families.
Our analysis examines this relationship within hierarchical fusion: we measure changes in spatial detail and class neighborhoods, together with the decoding response to latent perturbations.

\Needspace{0.46\textheight}
\section{HiRAE: controlled hierarchical composition}
\label{sec:method}
\begin{wrapfigure}[22]{r}{0.50\textwidth}
\vspace{-2\intextsep}
\centering
\includegraphics[width=\linewidth]{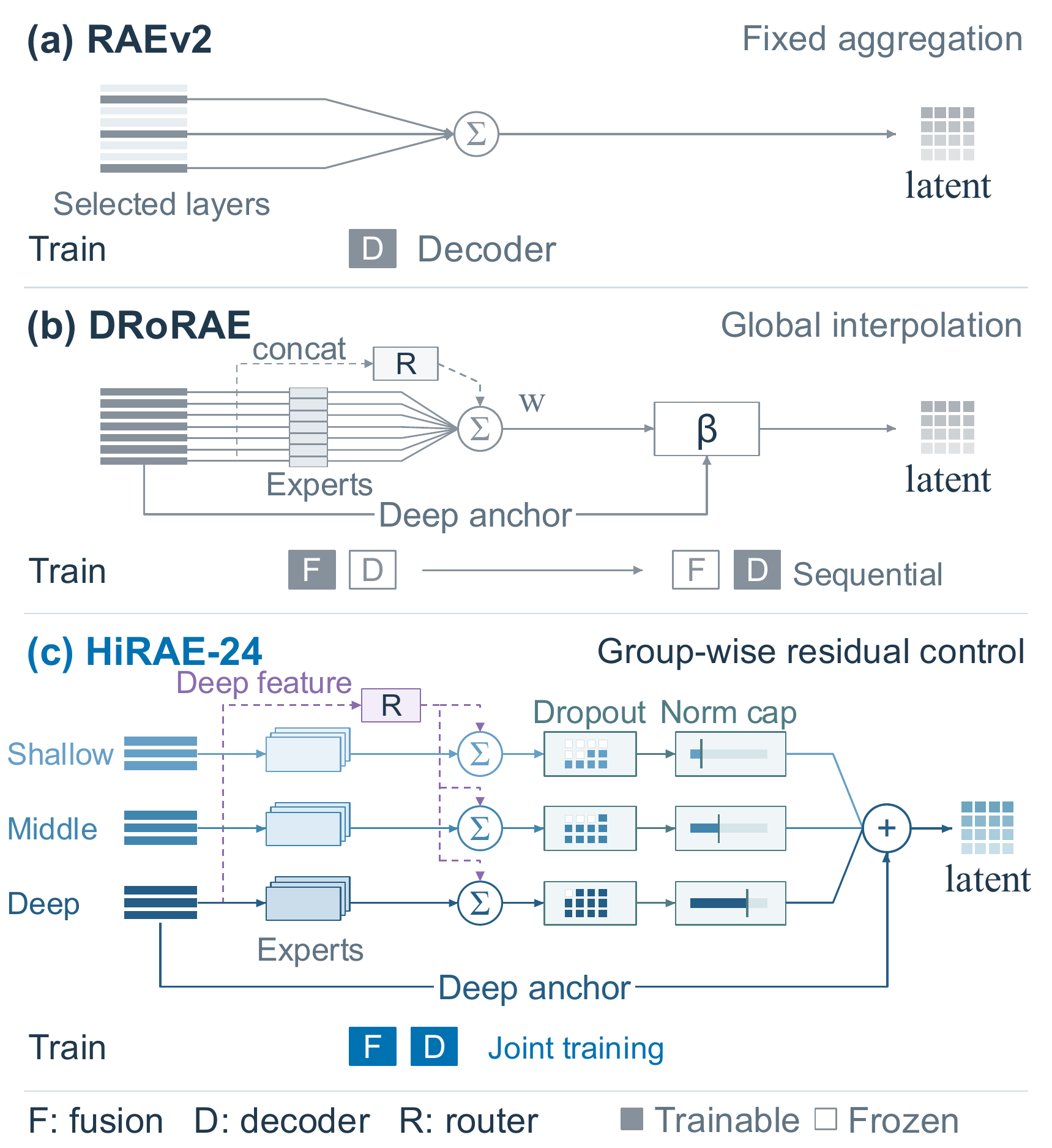}
\caption{\textbf{Layer use and tokenizer training} in RAEv2, DRoRAE, and HiRAE-24.}
\label{fig:transition}
\end{wrapfigure}

\begin{figure}[tb]
\centering
\includegraphics[width=0.9\linewidth]{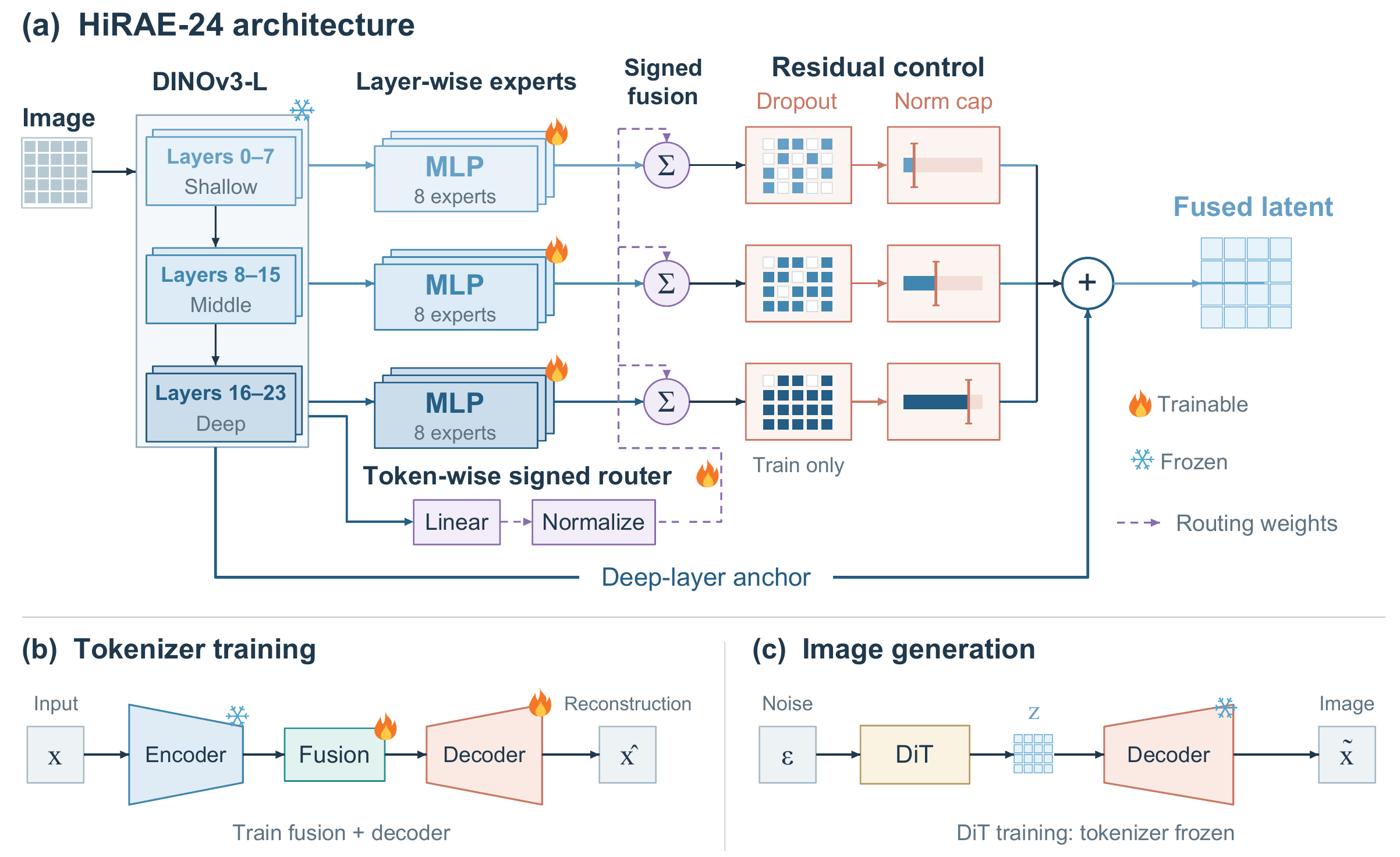}
\caption{\textbf{HiRAE-24 architecture and training.} (a) Layer-wise experts and signed routing combine all 24 encoder layers under depth-dependent residual controls. (b) Tokenizer training jointly updates fusion and decoder while freezing the encoder. (c) Generator training freezes the tokenizer.}
\label{fig:framework}
\end{figure}

\liftfull{} learns to use all encoder layers while jointly training the fusion module and decoder (Figure~\ref{fig:transition}). A separate learned transformation (expert) processes each layer's output, and a router learns the expert contributions at each spatial location. To prevent the latent space from drifting toward a reconstruction-dominated distribution during joint training, HiRAE combines these outputs into shallow, middle, and deep residual groups around the deepest-layer anchor. Group-wise norm caps assign tighter correction budgets to shallower groups, with residual dropout providing additional regularization. The encoder stays frozen, and the fused latent preserves its token count and channel dimension (Figure~\ref{fig:framework}).

\subsection{Layer-wise experts}
\label{sec:layer_experts}
A separate token-wise MLP expert~\citep{zhu2026drorae} transforms each frozen encoder feature $H_\ell\in\mathbb{R}^{N\times C}$ before fusion:
\begin{equation}
 U_\ell=\mathcal{E}_\ell(H_\ell),
 \qquad \ell=0,\ldots,L-1.
 \label{eq:expert}
\end{equation}
We use all $L=24$ DINOv3-L~\citep{simeoni2025dinov3} layers with $N=256$ tokens and $C=1024$ channels; expert implementation details are given in Appendix~\ref{app:fusion_modules}.

\subsection{Routing}
A shared linear projection of the deepest feature $H_{L-1}$ produces routing scores at each spatial token $n$. We apply $\ell_2$ normalization to these scores, retaining their signs:
\begin{equation}
 a_n=\operatorname{Linear}_{R}(H_{L-1,n})\in\mathbb{R}^{L},
 \qquad w_n=\frac{a_n}{\norm{a_n}_2}.
 \label{eq:router}
\end{equation}
Stacking $w_n$ gives $W\in\mathbb{R}^{N\times L}$, whose column $w_\ell=W_{:,\ell}$ weights layer $\ell$ at each spatial location. Normalization details are given in Appendix~\ref{app:fusion_modules}.

\subsection{Residual regularization}
\label{sec:residual_controls}
Routing controls the combination weights but does not directly bound the resulting feature correction. We retain $H_{L-1}$ as the anchor, but replace DRoRAE's global interpolation with \emph{residual regularization} applied separately to each depth group: groupwise norm caps and residual dropout. For the 24-layer encoder, we use three contiguous depth groups: $G_s=\{0,\ldots,7\}$, $G_m=\{8,\ldots,15\}$, and $G_d=\{16,\ldots,23\}$.

Each group forms an unregularized residual $R_g=\sum_{\ell\in G_g}w_\ell\odot U_\ell$, where $\odot$ broadcasts each spatial weight across channels. The residual-control module $\mathcal{C}_g$ converts $R_g$ into a controlled correction $\Delta_g$. We add these corrections to the deepest feature $H_{23}$ and apply layer normalization (LN) to obtain the fused latent $Z$:
\begin{equation}
 Z=\LN\!\left[H_{23}+\sum_{g\in\{s,m,d\}}\Delta_g\right],
 \qquad \Delta_g=\mathcal{C}_g(R_g;H_{23}),
 \label{eq:residual}
\end{equation}

Here, $\LN$ normalizes the $C$ channels of each spatial token independently. The module $\mathcal{C}_g$ first applies \emph{residual dropout} with probabilities $(p_s,p_m,p_d)=(0.50,0.25,0.10)$ during tokenizer training. It then scales down a group residual only when its norm exceeds its assigned budget. These \emph{groupwise norm caps} enforce
\begin{equation}
 \norm{\Delta_g}_F\leq c_g\norm{H_{23}}_F,
 \qquad (c_s,c_m,c_d)=(0.025,0.075,0.150).
 \label{eq:group_caps}
\end{equation}
Here, $\norm{\cdot}_F$ denotes the Frobenius norm, computed separately for each image over all spatial tokens and channels. The caps therefore bound the summed contribution of each depth group after routing. Shallower groups receive tighter norm budgets and stronger dropout, while middle and deep groups allow progressively larger corrections.

Together, the three group budgets bound the total correction before final normalization. By the triangle inequality,
\begin{equation}
 \norm{\Delta_s+\Delta_m+\Delta_d}_F
 \leq \norm{\Delta_s}_F+\norm{\Delta_m}_F+\norm{\Delta_d}_F
 \leq 0.250\,\norm{H_{23}}_F.
 \label{eq:total_residual_bound}
\end{equation}

\subsection{Training the tokenizer and generator}
\label{sec:training}
HiRAE jointly trains fusion and decoding within the tokenizer stage, then trains the generator on the frozen tokenizer's latents. In Stage 1, reconstruction losses update both the fusion module and decoder while the pretrained backbone stays frozen. DRoRAE~\citep{zhu2026drorae} instead trains fusion against a frozen decoder before decoder adaptation; HiRAE removes this separate fusion-only adaptation phase (Figure~\ref{fig:transition}). We use pixel reconstruction, perceptual, and adversarial losses, with decoder-input noise. In schematic form,
\begin{equation}
 \mathcal L_{\rm tok}=\mathcal L_{1}(x,\widehat x)
 +\lambda_{\rm perc}\mathcal L_{\rm perc}(x,\widehat x)
 +\lambda_{\rm adv}(e)\mathcal L_{\rm adv}(\widehat x).
\end{equation}
Here, $e$ denotes the training epoch. In Stage 2, we freeze the tokenizer and train a DiT generator on its latent representations, following RAEv2's prediction and internal-guidance framework. Appendix~\ref{app:implementation} provides the loss weighting and training configuration.

\section{Reconstruction and generation}
\label{sec:experiments}
\paragraph{Datasets.}
For reconstruction and class-conditional generation, we train and evaluate HiRAE on ImageNet-1K~\citep{deng2009imagenet}. For image reconstruction, we train the tokenizer on the training split at $256\times256$ resolution and evaluate on the validation split. Following the ADM evaluation protocol~\citep{dhariwal2021diffusion}, we generate 50,000 images per configuration for FID computation. For text-to-image (T2I) generation, we follow RAEv2~\citep{singh2026raev2} and pretrain on JourneyDB~\citep{sun2023journeydb} together with the long-caption and short-caption subsets of BLIP3o~\citep{chen2025blip3o}, using $256\times256$ images. We then apply supervised fine-tuning (SFT) on BLIP3o-60k.

\paragraph{Evaluation metrics.}
For ImageNet, we measure reconstruction quality with reconstruction FID (rFID), and generation quality with generation FID (gFID) and Inception Score (IS). On a matched reconstruction subset, we additionally measure peak signal-to-noise ratio (PSNR) and Learned Perceptual Image Patch Similarity (LPIPS; \citealp{zhang2018lpips}). We also report $\fdr$, which aggregates normalized Fr{\'e}chet distances across six representation spaces. Our evaluations use their arithmetic mean. For T2I, we evaluate text--image alignment with GenEval~\citep{ghosh2023geneval} and Dense Prompt Graph Benchmark (DPG-Bench; \citealp{hu2024ella}). We additionally report GenAI-Bench~\citep{li2024genaibench}, which evaluates compositional text--image alignment.

\paragraph{Implementation details.}
Our main comparisons evaluate \liftfull{}, which fuses all 24 layers of a frozen DINOv3-L encoder into a $16\times16\times1024$ latent. For ImageNet, we evaluate exponential moving average (EMA) generators with and without internal guidance, retaining class conditioning in both settings. Appendix~\ref{app:implementation} provides the ImageNet training and sampling configuration.
For T2I, we follow RAEv2. Pretraining uses 100K optimizer updates; SFT continues from the corresponding pretrained weights. Appendix~\ref{app:text_to_image} details training schedules and scoring protocols.

\begin{figure}[!t]
\centering
\includegraphics[width=\linewidth]{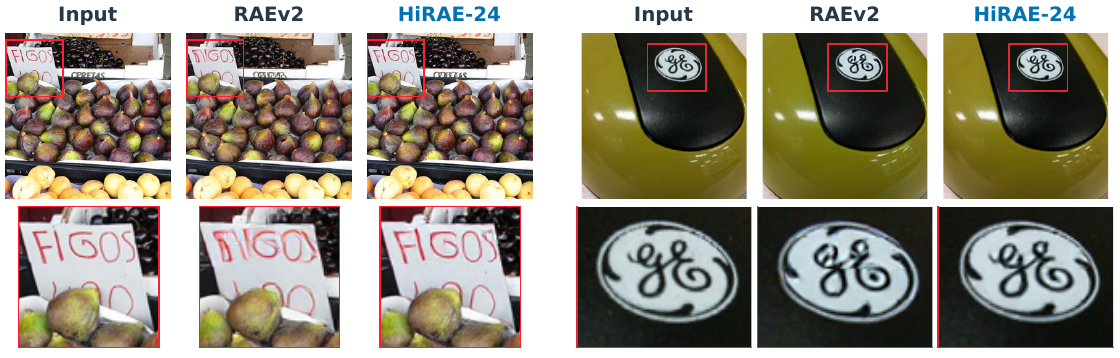}
\caption{\textbf{Matched reconstruction details.} Each selected triplet shows the input, official RAEv2, and HiRAE-24. Red boxes mark corresponding regions, enlarged below.}
\label{fig:reconstruction}
\end{figure}

\subsection{Reconstruction quality}
\label{sec:reconstruction}
\begin{table}[tb]
\centering
\caption{\textbf{ImageNet-256 reconstruction.} The upper block follows each source's evaluation protocol; dashes denote unreported values. The lower block combines 50K rFID with PSNR/LPIPS on our matched 5K subset (100 classes; AlexNet LPIPS). Bold marks the best result within the lower block.}
\label{tab:reconstruction}
\small
\begin{fullwidthtabular}{5}{llrrr}
\toprule
Tokenizer & Configuration & rFID $\downarrow$ & PSNR (dB) $\uparrow$ & LPIPS $\downarrow$\\
\midrule
SD-VAE~\citep{yao2025vavae} & f8, 4 channels & 0.610 & 26.90 & 0.130\\
VA-VAE~\citep{yao2025vavae} & f16, 32 channels & 0.280 & 27.96 & 0.096\\
REPA-E~\citep{leng2025repae} & VA-VAE + E2E tuning & 0.280 & 26.25 & 0.110\\
RPiAE~\citep{gong2026rpiae} & Pivot + variational bridge & 0.500 & 21.30 & 0.216\\
FAE~\citep{gao2025fae} & 32-channel feature AE & 0.680 & \textendash & \textendash\\
RAE~\citep{zheng2025rae} & DINOv2-B, last layer & 0.570 & 18.80 & 0.256\\
DRoRAE~\citep{zhu2026drorae} & DINOv2-B, three-phase & 0.290 & 24.32 & 0.134\\
DecQ~\citep{wang2026decq} & DINOv2-B + 8 queries & 0.470 & 22.76 & \textendash\\
HAE~\citep{chang2026hae} & DINOv3-L, spherical latent & 0.780 & 25.20 & \textendash\\
\midrule
RAEv2~\citep{singh2026raev2} & DINOv3-L, selected 7 layers & 0.299 & 22.667 & 0.074\\
\rowcolor{ourslight}\liftfull{} & DINOv3-L, all 24 layers & \textbf{0.209} & \textbf{26.377} & \textbf{0.043}\\
\bottomrule
\end{fullwidthtabular}

\par\smallskip
\end{table}

\textbf{HiRAE improves reconstruction fidelity within the original latent dimensions.} With the same frozen DINOv3-L encoder and $16\times16\times1024$ latent shape, \liftfull{} reduces rFID from the RAEv2 result of 0.299 to 0.209, a reduction of approximately 30\% (Table~\ref{tab:reconstruction}). On the matched 5,000-image subset, PSNR increases from 22.667 to 26.377 dB and LPIPS decreases from 0.074 to 0.043. The improvement therefore covers both pixel accuracy and perceptual similarity. Learned full-hierarchy fusion and joint decoder training recover finer image detail without increasing the generator's latent token count or channel dimension.
As shown in Figure~\ref{fig:reconstruction}, \liftfull{} more faithfully preserves text strokes and local colors. RAEv2 retains the overall image content but exhibits distortions in fine structures and local color shifts. These comparisons complement the rFID improvement, showing that controlled hierarchical fusion can recover image-specific details while maintaining the scene structure.
The improvement also extends across all four quartiles of original-image texture strength. More textured images show larger LPIPS reductions and a larger effect from removing the shallow residual group (Appendix~\ref{app:texture_strata}), complementing the visual examples.


\Needspace{9\baselineskip}
\subsection{Image generation}
\label{sec:generation}
\begin{table}[!t]
\centering
\caption{\textbf{Guided ImageNet-256 generation.} Epochs count generator training. CFG-int.: interval CFG; AG: AutoGuidance; IG: internal guidance.}
\label{tab:guided}
\small
\setlength{\tabcolsep}{4pt}
\begin{tabular}{lrlrrr}
\toprule
System & Epochs & Guide & gFID $\downarrow$ & IS $\uparrow$ & $\fdr$ $\downarrow$\\
\midrule
DiT-XL/2~\citep{peebles2023dit} & 1400 & CFG & 2.270 & 278.200 & ---\\
SiT-XL/2~\citep{ma2024sit} & 1400 & CFG & 2.060 & 270.300 & ---\\
REPA~\citep{yu2025repa} & 800 & CFG-int. & 1.420 & 305.700 & ---\\
VA-VAE~\citep{yao2025vavae} & 800 & CFG & 1.350 & 295.300 & ---\\
REPA-E~\citep{leng2025repae} & 800 & CFG & 1.120 & 302.900 & ---\\
RPiAE~\citep{gong2026rpiae} & 80 & AG & 1.510 & 225.900 & ---\\
FAE~\citep{gao2025fae} & 800 & CFG & 1.290 & 268.000 & ---\\
RAE~\citep{zheng2025rae} & 800 & AG & 1.130 & 262.600 & ---\\
RAE~\citep{zheng2025rae} & 80 & AG & 1.740 & 235.000 & ---\\
DRoRAE~\citep{zhu2026drorae} & 80 & AG & 1.650 & 230.600 & ---\\
LV-RAE~\citep{liu2026lvrae} & 800 & AG & 1.820 & 249.700 & ---\\
DecQ~\citep{wang2026decq} & 800 & AG & 1.050 & 259.600 & ---\\
HAE~\citep{chang2026hae} & 550 & CFG & 1.900 & 252.700 & ---\\
\midrule
\multicolumn{6}{l}{\textit{DINOv3-L / RAEv2 framework; internal guidance}}\\
RAEv2~\citep{singh2026raev2} & 80 & IG & 1.060 & 255.300 & 2.170\\
\rowcolor{ourslight}\liftfull{} & 80 & IG & \textbf{1.038} & \textbf{257.823} & \textbf{1.856}\\
\bottomrule
\end{tabular}
\end{table}

\begin{wraptable}[7]{R}{0.5\textwidth}
\vspace{-\intextsep}
\setlength{\abovecaptionskip}{0pt}
\centering
\caption{\textbf{Unguided ImageNet-256 generation.}}
\label{tab:unguided}
\small
\setlength{\tabcolsep}{3pt}
\begin{fullwidthtabular}{4}{lrrr}
\toprule
Tokenizer & gFID $\downarrow$ & IS $\uparrow$ & $\fdr$ $\downarrow$\\
\midrule
RAEv2 & 1.650 & 228.000 & 3.950\\
RAEv2 K=23 & 3.010 & 206.000 & ---\\
\midrule
\rowcolor{ourslight}\liftfull{} & 2.129 & 210.339 & 4.660\\
\bottomrule
\end{fullwidthtabular}
\end{wraptable}

\begin{figure}[tb]
\centering
\includegraphics[trim=0 182.4bp 0 0,clip,width=\linewidth]{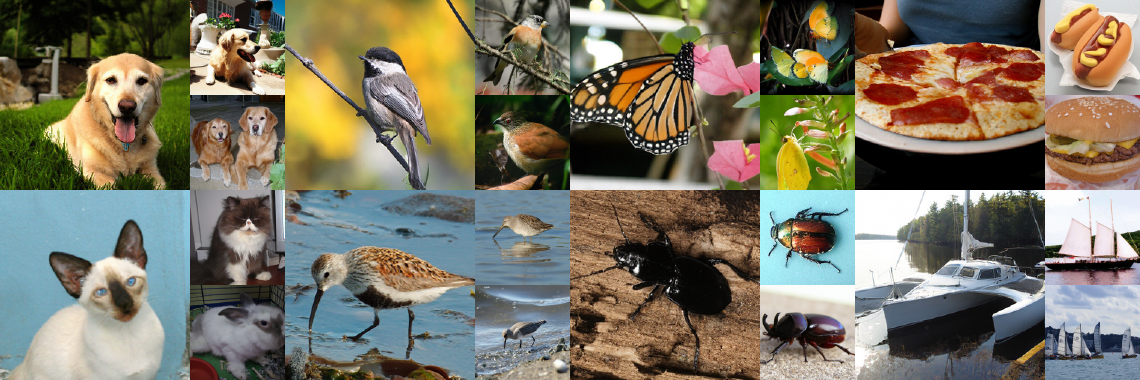}
\caption{\textbf{Guided samples from HiRAE-24.} Twenty-four selected class-conditioned images form eight visual groups, each with one larger example and two related samples.}
\label{fig:guided_samples}
\end{figure}

\textbf{Higher reconstruction fidelity coexists with competitive guided generation.}
As shown in Table~\ref{tab:guided}, \liftfull{} achieves a guided gFID of 1.038 after 80 epochs of generator training, compared with 1.060 for RAEv2 at the same training duration. IS also increases from 255.300 to 257.823. The reconstruction gain therefore coexists with competitive guided generation in the same representation learned from the full encoder hierarchy. DecQ~\citep{wang2026decq} appends eight detail-query tokens. It generates these alongside the original patch tokens, whereas \liftfull{} integrates hierarchical information into the existing patch-token layout. Their comparable gFID shows that detail enrichment can support competitive generation within the original latent token count and channel dimension. REPA-E~\citep{leng2025repae} obtains its tokenizer through end-to-end VAE--diffusion tuning; HiRAE learns fusion and decoding over a frozen encoder, then freezes the tokenizer for generator training. Figure~\ref{fig:guided_samples} shows selected outputs spanning animals, objects, and scenes, combining coherent object structure with fine local detail. Without guidance, Table~\ref{tab:unguided} shows that \liftfull{} achieves a gFID of 2.129, improving on RAEv2 K=23's 3.010 while remaining above RAEv2's 1.650.

\subsection{Text-to-image generation}
\label{sec:text_to_image}
HiRAE-7 and HiRAE-24 achieve higher text--image alignment scores than RAEv2 both after pretraining and after supervised fine-tuning (SFT). Table~\ref{tab:text_to_image} compares FLUX-VAE~\citep{labs2024flux}, RAEv2~\citep{singh2026raev2}, HiRAE-7, and \liftfull{} as frozen image tokenizers after 100K generator pretraining steps and after SFT. At inference, all four evaluated configurations use classifier-free guidance (CFG) with scale 6 and internal guidance disabled. We report GenEval, DPG-Bench, and GenAI-Bench scores; Appendix~\ref{app:text_to_image} gives implementation details.
After pretraining, \liftfull{} improves GenEval by 4.52 points and DPG-Bench by 1.51 points over RAEv2. This advantage persists after SFT, with gains of 2.84, 1.45, and 0.97 points on GenEval, DPG-Bench, and GenAI-Bench, respectively. HiRAE-7 also exceeds RAEv2 on every available benchmark, while \liftfull{} improves further across both stages. These results extend the evidence for learned fusion from class-conditioned generation to text-conditioned generation and support the full-depth configuration without prior layer-subset selection. Figure~\ref{fig:geneval_selected} in Appendix~\ref{app:qualitative_t2i} compares the four methods on five selected GenEval prompts after SFT.

\begin{table}[t]
\centering
\caption{\textbf{Text-to-image generation before and after SFT.} DPG and GenAI denote DPG-Bench and GenAI-Bench. Bold denotes the best available score within each stage.}
\label{tab:text_to_image}
\small
\begin{fullwidthtabular}{7}{lcccccc}
\toprule
 & \multicolumn{3}{c}{Pretraining} & \multicolumn{3}{c}{Finetuning}\\
\cmidrule(lr){2-4}\cmidrule(lr){5-7}
Model & GenEval $\uparrow$ & DPG $\uparrow$ & GenAI $\uparrow$ & GenEval $\uparrow$ & DPG $\uparrow$ & GenAI $\uparrow$\\
\midrule
FLUX-VAE~\citep{labs2024flux} & 49.73& 78.86& 63.65& 84.13& 83.48& 68.75\\
RAEv2~\citep{singh2026raev2} & 56.42 & 81.22 & 67.02 & 84.86 & 84.90 & 71.69\\
HiRAE-7 & 58.77 & 81.60 & 67.13 & 85.17 & 85.05 & 72.02\\
\rowcolor{ourslight}\liftfull{} & \textbf{60.94} & \textbf{82.73} & \textbf{68.08} & \textbf{87.70} & \textbf{86.35} & \textbf{72.66}\\
\bottomrule
\end{fullwidthtabular}
\end{table}

\section{Ablation Studies and Analysis}
\label{sec:discussion}

\subsection{Ablation studies}
\label{sec:fusion_designs}

\begin{table}[t]
\centering
\caption{\textbf{Expert inputs and residual regularization.} Raw-layer experts: $\checkmark$, 24 layer experts; $\times$, seven depth-mode experts. Residual regularization combines norm caps and residual dropout.}
\label{tab:ablations}
\small
\begin{fullwidthtabular}{8}{lccrrrrr}
\toprule
\multirow{2}{*}{Model} & \multirow{2}{*}{\shortstack{Raw-layer\\experts}} & \multirow{2}{*}{\shortstack{Residual\\regularization}} & \multirow{2}{*}{rFID $\downarrow$} & \multicolumn{2}{c}{20 epochs} & \multicolumn{2}{c}{80 epochs}\\
\cmidrule(lr){5-6}\cmidrule(lr){7-8}
 & & & & gFID $\downarrow$ & $\fdr$ $\downarrow$ & gFID $\downarrow$ & $\fdr$ $\downarrow$\\
\midrule
\rowcolor{ourslight}\lift{} & $\checkmark$ & $\checkmark$ & 0.209 & \textbf{2.242} & \textbf{2.533} & \textbf{1.038} & \textbf{1.856}\\
\lift{} & $\times$ & $\checkmark$ & 0.230 & 2.410 & 2.650 & 1.067 & 1.929\\
\lift{} & $\times$ & $\times$ & \textbf{0.023} & 7.905 & 14.722 & \textendash & \textendash\\
\bottomrule
\end{fullwidthtabular}
\end{table}

\begin{wraptable}[13]{R}{0.5\textwidth}
\setlength{\abovecaptionskip}{0pt}
\centering
\caption{\textbf{Fusion methods in the RAEv2 framework.} DRoRAE-style fusion uses global interpolation to mix the deepest-layer representation and aggregated expert output at a fixed 80:20 ratio.}
\label{tab:full_depth_composition}
\small
\begin{fullwidthtabular}{4}{lrrr}
\toprule
Model & rFID $\downarrow$ & gFID $\downarrow$ & $\fdr$ $\downarrow$\\
\midrule
\multicolumn{4}{l}{\textit{Selected 7 layers}}\\
RAEv2 & 0.299 & 1.060 & 2.170\\
HiRAE-7 & 0.217 & \textbf{1.038} & 1.913\\
\midrule
\multicolumn{4}{l}{\textit{All 24 layers}}\\
\shortstack[l]{RAEv2 + DRoRAE} & \textbf{0.065} & 1.551 & 3.375\\
\rowcolor{ourslight}\liftfull{} & 0.209 & \textbf{1.038} & \textbf{1.856}\\
\bottomrule
\end{fullwidthtabular}
\end{wraptable}

\paragraph{Comparison of fusion methods.}
Applying DRoRAE-style layer experts and learned aggregation~\citep{zhu2026drorae} within RAEv2 improves both reconstruction and guided generation (Table~\ref{tab:full_depth_composition}). On the same seven selected layers, HiRAE-7 reduces rFID from 0.299 to 0.217 and guided gFID from 1.060 to 1.038. HiRAE-24 uses all 24 layers to remove the prerequisite of selecting a suitable subset, while HiRAE-7 remains a compact extension when a subset is available. For full-depth fusion, we compare HiRAE-24 with DRoRAE-style fusion adapted to RAEv2, using 24 experts in both configurations. The adapted DRoRAE-style fusion achieves lower rFID (0.065 versus 0.209), whereas HiRAE-24 achieves lower guided gFID (1.038 versus 1.551) and $\fdr$ (1.856 versus 3.375). These results show that reconstruction fidelity alone is insufficient for choosing a fusion method for guided generation. HiRAE-24 improves reconstruction over RAEv2 while achieving competitive guided generation performance, supporting its use for both reconstruction and generation.

\paragraph{Expert input and residual regularization ablation.}
Table~\ref{tab:ablations} compares three configurations trained for 16 tokenizer epochs, with generation evaluated on 50K guided samples from EMA generators at epochs 20 and 80. With residual regularization fixed, replacing 24 raw-layer experts with seven depth-mode experts increases rFID from 0.209 to 0.230 and guided gFID from 1.038 to 1.067 at epoch 80. The same ordering holds at epoch 20, supporting retention of the layer-wise expert design. The unregularized depth-mode configuration reaches rFID 0.023, but its guided gFID and $\fdr$ rise to 7.905 and 14.722 at epoch 20, compared with 2.410 and 2.650 for the regularized configuration. Together, these comparisons favor retaining layer-wise experts and residual regularization to combine reconstruction fidelity with guided generation quality.

\paragraph{Depth-group ablation.}
We compare two, three, and four depth groups. Three groups achieve the lowest guided gFID, with rFID close to that of two groups (Table~\ref{tab:group_count_results}). This balance supports our three-group design. In the main HiRAE-24 tokenizer, removing any depth-group residual increases reconstruction LPIPS on 5,000 matched images (Table~\ref{tab:main_interventions}), showing that all three groups contribute to reconstruction. For equal-norm removals, high-frequency removal causes more damage in the shallow group, while low-frequency removal causes more damage in the deep group. The middle group's difference is small, with a paired 95\% interval that includes zero. These results support complementary reconstruction information across depths.
We provide full results in Apps~\ref{app:group_count} and~\ref{app:group_interventions}.

\begin{table}[tb]
\begin{minipage}[t]{0.48\textwidth}
\vspace{0pt}
\centering
\caption{\textbf{Depth-group count.} rFID: 5K images; guided gFID: 10K samples, IG=1.78.}
\label{tab:group_count_results}
\small
\begin{fullwidthtabular}{3}{rrr}
\toprule
Groups & rFID $\downarrow$ & Guided gFID $\downarrow$\\
\midrule
2 & \textbf{3.293} & 29.385\\
\rowcolor{ourslight}3 & 3.363 & \textbf{28.354}\\
4 & 3.457 & 31.321\\
\bottomrule
\end{fullwidthtabular}
\end{minipage}
\hfill
\begin{minipage}[t]{0.48\textwidth}
\vspace{0pt}
\centering
\caption{\textbf{Depth-group interventions.}
LPIPS changes $\times10^3$.}
\label{tab:main_interventions}
\small
\begin{fullwidthtabular}{3}{lrr}
\toprule
Depth group & \shortstack{Group removal} & High $-$ low\\
\midrule
Shallow (0--7) & 5.650 & 0.215\\
Middle (8--15) & 54.111 & 0.068\\
Deep (16--23) & 147.750 & $-2.053$\\
\bottomrule
\end{fullwidthtabular}
\end{minipage}
\end{table}

\subsection{Latent-space analysis}
\label{sec:latent_analysis}
\begin{figure}[t]
\centering
\includegraphics[width=0.90\linewidth]{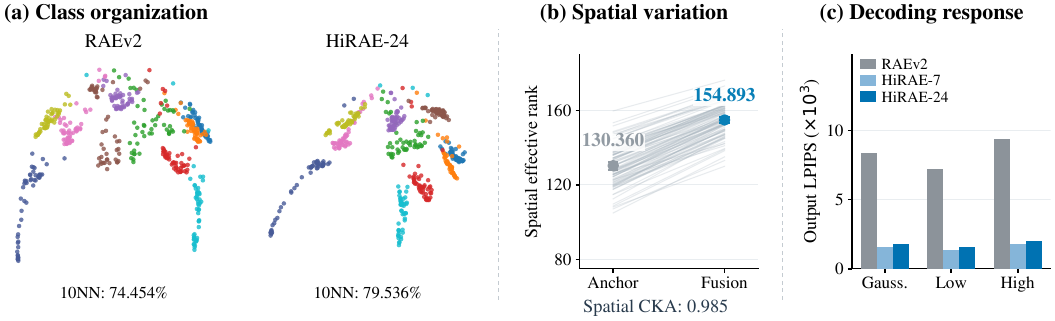}
\caption{\textbf{Latent structure and decoding sensitivity.} (a) Independently fitted PHATE views of the same images; colors denote classes. Same-class 10NN uses the original feature space. (b) Within HiRAE-24, lines connect anchor--fusion effective ranks for 100 images; points and error bars show means and 95\% bootstrap CIs. (c) Mean output LPIPS under 10\% relative latent perturbations.}
\label{fig:latent_analysis}
\end{figure}

\textbf{Fusion expands spatial variation while retaining class organization.} Figure~\ref{fig:latent_analysis}(a) visualizes RAEv2 and \liftfull{} on the same 500 images using PHATE~\citep{moon2019phate}, which provides a low-dimensional visualization of the representations. In the original feature space, the same-class fraction among ten nearest neighbors over 5,000 matched images is 74.454\% for RAEv2 and 79.536\% for \liftfull{}.
Within \liftfull{}, this fraction changes from 79.634\% for the deep anchor to 79.536\% after fusion, indicating largely preserved class neighborhoods. Across 100 fixed images, one per class, spatial effective rank increases from 130.360 for the deep anchor, $\LN(H_{23})$, to 154.893 for the fused representation, with an increase in every image (Figure~\ref{fig:latent_analysis}(b)). Meanwhile, mean spatial centered kernel alignment (CKA; \citealp{kornblith2019similarity}) remains 0.985. Thus, spatial variation spreads over more feature directions while the patch-relationship structure remains similar.

\textbf{Spatial enrichment coexists with lower decoding sensitivity.} Figure~\ref{fig:latent_analysis}(c) measures LPIPS between clean and perturbed decodings under Gaussian, low-frequency, and high-frequency latent perturbations. We measure each tokenizer's decoding response through its own inverse normalization and trained decoder. At a perturbation norm equal to 10\% of the standardized latent norm, \liftfull{} produces 21\%--22\% of RAEv2's output LPIPS change across the three direction types.
The trained generator also predicts \liftfull{} latents more accurately across intermediate noise levels. At seven matched coefficient log signal-to-noise ratio (log-SNR) levels, \liftfull{} has lower clean-latent mean squared error than RAEv2 at the five interior levels and slightly higher error at both endpoints. Appendix~\ref{app:mechanism} provides the full results and protocol. The diagnostics show lower latent prediction error at intermediate noise levels and lower decoding sensitivity to the perturbations.

\enlargethispage{2\baselineskip}
\section{Conclusion}
\label{sec:conclusion}
HiRAE learns high-fidelity visual tokenizers through joint fusion and decoder training, with depth-dependent residual bounds controlling each depth group's contribution. HiRAE-24 learns the contributions of all 24 encoder layers without manual layer-subset selection or a separate fusion-only adaptation phase. ImageNet-256 results show higher reconstruction fidelity than RAEv2 with competitive guided generation within the original latent dimensions. Matched-image analyses connect the added detail to complementary information across depth groups and largely preserved class neighborhoods. HiRAE-7 extends the framework to compact fusion over an established subset.

\FloatBarrier
\label{maintextend}
\clearpage
\bibliography{references}
\bibliographystyle{plainnat}
\clearpage
\appendix
\raggedbottom
\section*{Appendix Guide}
\begingroup
\small
\setlength{\parskip}{1pt}
\textbf{\hyperref[app:implementation]{\ref*{app:implementation} Implementation Details}}\nobreak\dotfill\pageref{app:implementation}\par
\hspace*{1em}\hyperref[app:architecture]{\ref*{app:architecture} HiRAE-24 architecture}\nobreak\dotfill\pageref{app:architecture}\par
\hspace*{1em}\hyperref[app:imagenet_training]{\ref*{app:imagenet_training} ImageNet training}\nobreak\dotfill\pageref{app:imagenet_training}\par
\hspace*{1em}\hyperref[app:imagenet_sampling]{\ref*{app:imagenet_sampling} ImageNet sampling}\nobreak\dotfill\pageref{app:imagenet_sampling}\par
\hspace*{1em}\hyperref[app:text_to_image]{\ref*{app:text_to_image} Text-to-image training and evaluation}\nobreak\dotfill\pageref{app:text_to_image}\par
\textbf{\hyperref[app:metrics]{\ref*{app:metrics} Evaluation Protocols and Baseline Results}}\nobreak\dotfill\pageref{app:metrics}\par
\hspace*{1em}\hyperref[app:metric_definitions]{\ref*{app:metric_definitions} Metrics and aggregation}\nobreak\dotfill\pageref{app:metric_definitions}\par
\hspace*{1em}\hyperref[app:comparison_sources]{\ref*{app:comparison_sources} Baseline configurations and result sources}\nobreak\dotfill\pageref{app:comparison_sources}\par
\hspace*{1em}\hyperref[app:official_unguided]{\ref*{app:official_unguided} Official-checkpoint unguided evaluation}\nobreak\dotfill\pageref{app:official_unguided}\par
\hspace*{1em}\hyperref[app:additional_generation]{\ref*{app:additional_generation} Additional generation comparisons}\nobreak\dotfill\pageref{app:additional_generation}\par
\textbf{\hyperref[app:additional_experiments]{\ref*{app:additional_experiments} Additional Reconstruction Results}}\nobreak\dotfill\pageref{app:additional_experiments}\par
\hspace*{1em}\hyperref[app:paired_reconstruction]{\ref*{app:paired_reconstruction} Matched-image evaluation}\nobreak\dotfill\pageref{app:paired_reconstruction}\par
\hspace*{1em}\hyperref[app:reconstruction_detail]{\ref*{app:reconstruction_detail} Paired improvements and spatial-detail errors}\nobreak\dotfill\pageref{app:reconstruction_detail}\par
\hspace*{1em}\hyperref[app:texture_strata]{\ref*{app:texture_strata} Reconstruction gains across texture strata}\nobreak\dotfill\pageref{app:texture_strata}\par
\textbf{\hyperref[app:design]{\ref*{app:design} Fusion Configurations and Ablations}}\nobreak\dotfill\pageref{app:design}\par
\hspace*{1em}\hyperref[app:lifae7]{\ref*{app:lifae7} HiRAE-7: selected-layer extension}\nobreak\dotfill\pageref{app:lifae7}\par
\hspace*{1em}\hyperref[app:global_interpolation]{\ref*{app:global_interpolation} DRoRAE-style full-depth fusion}\nobreak\dotfill\pageref{app:global_interpolation}\par
\hspace*{1em}\hyperref[app:expert_variants]{\ref*{app:expert_variants} Expert inputs and residual regularization}\nobreak\dotfill\pageref{app:expert_variants}\par
\hspace*{1em}\hyperref[app:group_count]{\ref*{app:group_count} Number of depth groups}\nobreak\dotfill\pageref{app:group_count}\par
\hspace*{1em}\hyperref[app:group_interventions]{\ref*{app:group_interventions} Layer-group and frequency interventions}\nobreak\dotfill\pageref{app:group_interventions}\par
\textbf{\hyperref[app:mechanism]{\ref*{app:mechanism} Latent-Space and Decoding Analysis}}\nobreak\dotfill\pageref{app:mechanism}\par
\hspace*{1em}\hyperref[app:analysis_protocol]{\ref*{app:analysis_protocol} Analysis protocol}\nobreak\dotfill\pageref{app:analysis_protocol}\par
\hspace*{1em}\hyperref[app:feature_geometry]{\ref*{app:feature_geometry} Class organization and spatial enrichment}\nobreak\dotfill\pageref{app:feature_geometry}\par
\hspace*{1em}\hyperref[app:spatial_comparison]{\ref*{app:spatial_comparison} Cross-system spatial statistics}\nobreak\dotfill\pageref{app:spatial_comparison}\par
\hspace*{1em}\hyperref[app:decoding_response]{\ref*{app:decoding_response} Decoding response to latent perturbations}\nobreak\dotfill\pageref{app:decoding_response}\par
\hspace*{1em}\hyperref[app:generator_error]{\ref*{app:generator_error} Generator prediction error}\nobreak\dotfill\pageref{app:generator_error}\par
\textbf{\hyperref[app:qualitative]{\ref*{app:qualitative} Additional Qualitative Results}}\nobreak\dotfill\pageref{app:qualitative}\par
\hspace*{1em}\hyperref[app:qualitative_reconstruction]{\ref*{app:qualitative_reconstruction} Reconstruction comparisons}\nobreak\dotfill\pageref{app:qualitative_reconstruction}\par
\hspace*{1em}\hyperref[app:qualitative_generation]{\ref*{app:qualitative_generation} Generation samples}\nobreak\dotfill\pageref{app:qualitative_generation}\par
\hspace*{1em}\hyperref[app:sample_selection]{\ref*{app:sample_selection} Sample selection and visualization details}\nobreak\dotfill\pageref{app:sample_selection}\par
\hspace*{1em}\hyperref[app:qualitative_t2i]{\ref*{app:qualitative_t2i} Text-to-image comparisons}\nobreak\dotfill\pageref{app:qualitative_t2i}\par
\endgroup
\clearpage
\section{Implementation Details}
\label{app:implementation}
\suppressfloats[t]
\subsection{HiRAE-24 architecture}
\label{app:architecture}
\suppressfloats[t]
The backbone is DINOv3 ViT-L/16, with the LVD-1689M pretrained weights. We extract patch tokens from all 24 blocks using the encoder's normalized intermediate-layer interface. These tensors constitute $H_\ell$ in Section~\ref{sec:method}. There is no discrete cosine transform (DCT) or other depth compression in \liftfull{}. The spatial resolution is $16\times16$ at input resolution $256\times256$. Each of the 24 experts has linear dimensions $1024\rightarrow4096\rightarrow1024$, with hidden LayerNorm and GELU. The expert's internal dropout probability is zero in the main configuration; the group residual dropout is distinct.

\paragraph{Fusion modules.}
\label{app:fusion_modules}
The three modules in Section~\ref{sec:method} collect the following operations. Let $\LN$ denote token-wise channel normalization and $\LN_h$ the expert's hidden normalization. A layer-specific expert is
\begin{equation}
 \begin{aligned}
 \mathcal{E}_\ell(H)&=\LN\!\left(E_\ell(\LN(H))\right),\\
 E_\ell(u)&=W_{\ell,2}\,\operatorname{GELU}\!\left(\LN_h(W_{\ell,1}u+b_{\ell,1})\right)+b_{\ell,2}.
 \end{aligned}
\end{equation}
The hidden width is $4C$. The router acts independently at each spatial token $n$:
\begin{equation}
 a_n=W_R H_{L-1,n}+b_R,\qquad
 W_{n,\ell}=\frac{a_{n,\ell}}{\sqrt{\max(\sum_{j=0}^{L-1}a_{n,j}^{2},\epsilon)}}.
\end{equation}
For the group residual $R_g=\sum_{\ell\in G_g}w_\ell\odot U_\ell$, define $D_g=\operatorname{Drop}_{p_g}(R_g)$. The residual-control module is
\begin{equation}
 \mathcal{C}_g(R_g;H_{L-1})=D_g
 \min\!\left(1,\frac{c_g\norm{H_{L-1}}_F}{\max(\norm{D_g}_F,\epsilon)}\right).
\end{equation}
Dropout is elementwise and disabled at inference. Norms cover all spatial tokens and channels within each image. The three groups contain layers 0--7, 8--15, and 16--23. Their caps sum to 0.250, so the triangle inequality gives
\begin{equation}
 \left\lVert\sum_g\mathcal{C}_g(R_g;H_{L-1})\right\rVert_F
 \leq\sum_gc_g\norm{H_{L-1}}_F=0.250\norm{H_{L-1}}_F.
\end{equation}
This bound holds before the final normalization. Following Equation~\ref{eq:residual}, the final latent is
\begin{equation}
 Z=\LN\!\left[H_{23}+\sum_{g\in\{s,m,d\}}\Delta_g\right],
 \qquad \Delta_g=\mathcal{C}_g(R_g;H_{23}).
\end{equation}

\begin{table}[!htbp]
\centering
\caption{\textbf{Residual-control settings for \liftfull{}.} Layer ranges use zero-based indices.}
\begin{tabular}{lrrr}
\toprule
Group & Layers & Norm cap & Dropout\\
\midrule
Shallow & 0--7 & 0.025 & 0.50\\
Middle & 8--15 & 0.075 & 0.25\\
Deep & 16--23 & 0.150 & 0.10\\
\bottomrule
\end{tabular}
\end{table}

We compute each group's residual cap separately for each image over all tokens and channels. We apply dropout before the cap, then add the controlled residuals to the deepest anchor before final normalization. At inference, we disable dropout and retain the norm caps.

\paragraph{Layer-count convention.}
Using zero-based indices, the released RAEv2 K=23 configuration selects blocks $1,\ldots,23$ of the 24-block DINOv3-L encoder, omitting block 0. Our full-depth configuration uses blocks $0,\ldots,23$, including block 0. These counts refer to Transformer block outputs; neither counts the patch embedding as an extra layer. The released K=23 encoder applies fixed aggregation: it averages normalized selected-layer patch features and adds the spatial mean of the deepest feature. We learn the expert transformations and routing weights, and apply depth-dependent residual controls. The released configuration is available in the \href{https://github.com/nanovisionx/RAEv2/blob/main/configs/stage2/training/imagenet-dinov3l-k23.yaml}{official repository}.

\subsection{ImageNet training}
\label{app:imagenet_training}
\suppressfloats[t]
\begin{table}[H]
\centering
\caption{\textbf{Training settings for \liftfull{}.}}
\small
\begin{fullwidthtabular}{3}{lll}
\toprule
Setting & Stage 1 & Stage 2\\
\midrule
Epochs & 16 & 80\\
Global batch size & 128 & 1024\\
Gradient accumulation & 1 & 2\\
EMA decay & 0.9978 & 0.9995\\
Peak learning rate & $2\times10^{-4}$ & $2\times10^{-4}$\\
Final learning rate & $2\times10^{-5}$ & $2\times10^{-5}$\\
LR schedule & Cosine, end epoch 16 & Linear, end epoch 50\\
Warmup & 1 epoch & 25 epochs\\
Weight decay & 0 & 0\\
Gradient clipping & Disabled & Norm 1.0\\
Decoder-input noise $\tau$ & 0.8 & 0\\
Frozen modules & DINOv3 backbone & Full tokenizer\\
\bottomrule
\end{fullwidthtabular}
\end{table}

Stage 1 uses the ViT-XL decoder configuration. Pixel reconstruction is an L1 loss. The perceptual loss~\citep{zhang2018lpips} has weight 1.0. The adversarial term has base weight 0.75, multiplied by the adaptive weight derived from decoder gradients, capped at 10000. Discriminator updates begin at epoch 6, and the decoder's adversarial objective begins at epoch 8. The DINO-based discriminator uses a hinge loss; the generator objective is the negative mean discriminator logit. These objectives train the decoder and fusion jointly. The tokenizer optimizer is AdamW with betas $(0.9,0.95)$.

The diffusion generator uses a DiT-with-DDT-head architecture with hidden widths $(1440,2048)$, depths $(28,2)$, attention-head counts $(20,16)$, MLP ratio 4, and an intermediate/base depth of 8. Its latent patch size is 1. Class conditioning uses eight class tokens and four time tokens; label-dropout probability is 0.1. The configuration specifies $x$ prediction and logit-normal time sampling, with time-distribution shift dimensions 262144 and base 4096. Stage 2 uses the configured GMuon optimizer, momentum 0.95 and Nesterov acceleration. The tokenizer's final EMA latent mean and variance are estimated before diffusion training.

\subsection{ImageNet sampling}
\label{app:imagenet_sampling}
\suppressfloats[t]
ImageNet generation metrics use 50,000 class-conditioned samples at $256\times256$, an epoch-80 EMA model, BF16, and 100 Euler ODE steps. Guided evaluation uses IG=1.78 on $[0.1,1.0]$, with CFG=1. Unguided evaluation sets both scales to 1 and retains the class input. We shuffle the generated samples with seed 0 before computing split-based Inception Score.

We generate HiRAE-24 unguided samples on eight GPUs with batch size 32 per GPU and HiRAE-7 samples on four GPUs with the same per-GPU batch size. We map global seed 42 to rank seeds $42\times\text{world size}+\text{rank}$. HiRAE-24 guided evaluation uses eight GPUs with batch size 2 per GPU. Each configuration uses its own generated sample set.

\subsection{Text-to-image training and evaluation}
\label{app:text_to_image}
\suppressfloats[t]
\paragraph{Tokenizers and generator.}
RAEv2, HiRAE-7, and HiRAE-24 use DINOv3-L/16 and produce a $16\times16$ grid of 1024-dimensional latent tokens. HiRAE-24 and HiRAE-7 use their epoch-16 EMA tokenizers. HiRAE-7 and RAEv2 use encoder layers $\{11,13,15,17,19,21,23\}$ with zero-based indexing. Our RAEv2 baseline uses the official DINOv3-L tokenizer, decoder, and latent statistics; we train its T2I generator under the same recipe as the HiRAE generators. FLUX-VAE uses the FLUX autoencoder paired with our trained DiT. Each configuration uses its own decoder and latent normalization statistics. Both T2I pretraining and SFT train only the generator, with the tokenizer, decoder, and text encoder frozen.

The DiT with DDT head follows RAEv2~\citep{singh2026raev2}, with backbone/head depths of 28/2, hidden widths of 1440/2048, and attention-head counts of 20/16. The MLP ratio is 4, and latent patch size is 1. The generator contains approximately 875M parameters, excluding the frozen modules. Qwen3-0.6B provides up to 256 caption tokens; the model also uses four time-conditioning tokens and condition dropout of 0.1. Training uses $x$ prediction, logit-normal time sampling with shift 8, and a time-denominator floor of 0.05. The transport objective converts the prediction to velocity and computes squared error. The internal-guidance base branch has depth 8 and loss coefficient 1.0. We disable REPA and additional tokenizer noise during generator training.

\paragraph{Data preparation.}
Pretraining streams 5,141 tar shards through WebDataset: 419 from JourneyDB, 2,891 from BLIP3o Long-Caption, and 1,831 from BLIP3o Short-Caption. We shuffle the combined shard list before distributing it across ranks and workers, with additional shard and sample shuffle buffers. Images undergo RGB conversion, bicubic resizing of the shorter edge to 256, and a $256\times256$ center crop; the loader skips decoding failures. SFT uses 58,859 decoded image--text pairs from 11 BLIP3o-60k shards. We cache the same preprocessing as lossless PNG images in an Arrow dataset and shuffle the dataset with a distributed sampler each epoch. Each SFT epoch contains 57 optimizer updates, covering 58,368 sample presentations after dropping incomplete accumulation groups.

\paragraph{Optimization.}
Table~\ref{tab:t2i_training} summarizes the shared training settings. Two-dimensional parameters use GMuon with momentum 0.95, Nesterov updates, and RMS-norm learning-rate adjustment; other parameters use AdamW with betas $(0.9,0.95)$ and $\epsilon=10^{-8}$. Both parameter groups use zero weight decay and the same learning-rate schedule. Pretraining holds the learning rate at $2\times10^{-4}$ for 50K updates, then follows a linear decay targeting $2\times10^{-5}$ at 150K updates. We evaluate at 100K updates, where the nominal learning rate is $1.1\times10^{-4}$. SFT initializes model and EMA weights from the corresponding 100K checkpoint and resets the optimizer and scheduler. It warms up for 100 updates and then decays linearly to $2\times10^{-5}$ over a total budget of 2,850 updates.

\begin{table}[!htbp]
\centering
\caption{\textbf{Text-to-image training settings.} Both phases train the same generator architecture with frozen visual and text encoders.}
\label{tab:t2i_training}
\small
\begin{fullwidthtabular}{3}{lll}
\toprule
Setting & Pretraining & SFT\\
\midrule
Hardware & $8\times$ H800 80GB & $8\times$ H800 80GB\\
Precision & BF16 mixed precision & BF16 mixed precision\\
Microbatch per GPU & 32 & 32\\
Gradient accumulation & 4 & 4\\
Global batch size & 1024 & 1024\\
Optimizer updates & 100,000 & 2,850\\
Peak learning rate & $2\times10^{-4}$ & $2\times10^{-4}$\\
Gradient clipping & Norm 1.0 & Norm 1.0\\
EMA decay & 0.9995 & 0.9995\\
Configuration seed & 42 & 42\\
\bottomrule
\end{fullwidthtabular}
\end{table}

\paragraph{Sampling and benchmark scoring.}
All reported T2I evaluations use EMA weights, BF16, 50 Euler steps, time shift 8, and CFG=6 over the full sampling interval. We disable internal guidance at inference while retaining its base-branch loss during training. We generate one $256\times256$ image per prompt: 553 images for GenEval, 1,065 for DPG-Bench, and 1,600 for GenAI-Bench-1600. All reported scores multiply the benchmark mean by 100.

For GenEval, we use the evaluation implementation released with RAEv2~\citep{singh2026raev2}. DPG-Bench uses \texttt{dpg-evaluator==0.1.0} with mPLUG VQA and question-dependency corrections. We first average question scores within each image and then average across images. GenAI-Bench uses CLIP-FlanT5-XL VQAScore averaged over paired images and prompts; this score is a continuous alignment measure. The scorer uses revision \texttt{3b4a6b1b618f4e286f5353b5b5147a3ae7d9ec55}.

GenEval and DPG-Bench evaluations use the training process's random-number state. GenAI-Bench sampling uses eight GPUs with 16 images per GPU and seed 42, assigning rank seeds as $42\times\text{world size}+\text{rank}$. GenEval and DPG-Bench retain the evaluation RNG state across checkpoints.

\FloatBarrier
\section{Evaluation Protocols and Baseline Results}
\label{app:metrics}
\suppressfloats[t]
\subsection{Metrics and aggregation}
\label{app:metric_definitions}
\suppressfloats[t]
We report Fr\'echet distances in the Inception, ConvNeXt, DINOv2, MAE, SigLIP, and CLIP representation spaces, following the multi-representation evaluation perspective of \citet{yang2026fdloss}. For normalized distances $d_j$, the aggregate used throughout our main tables is
\begin{equation}
 \mathrm{FD}_{r,\mathrm{arithmetic}}^6=\frac{1}{6}\sum_{j=1}^6 d_j.
\end{equation}
We report RAEv2's guided $\fdr$ of 2.170 from its Table 7~\citep{singh2026raev2}.

ImageNet results use three decimal places, and text-to-image benchmark scores use two decimal places. We compute aggregates and relative changes before rounding.

\subsection{Baseline configurations and result sources}
\label{app:comparison_sources}
\suppressfloats[t]
Tables~\ref{tab:reconstruction} and~\ref{tab:guided} broaden the comparison to established latent diffusion systems and recent representation-based tokenizers. Table~\ref{tab:unguided} focuses on the RAEv2 family. External entries reproduce the measurements reported in the cited papers; HiRAE entries are our evaluations. The upper blocks compare complete systems, and the lower blocks compare the RAEv2 family. RAEv2 reconstruction uses its ImageNet-only Table 14, generation gFID/IS uses Table 16, and guided $\fdr$ uses Table 7~\citep{singh2026raev2}.

\begin{table}[!htbp]
\centering
\caption{\textbf{Generator configurations for the main-table baselines.} Parameter counts exclude tokenizers and separate guidance models.}
\small
\begin{fullwidthtabular}{3}{llr}
\toprule
System & Generator & Parameters (M)\\
\midrule
DiT / SD-VAE & DiT-XL/2 & 675\\
SiT / SD-VAE & SiT-XL/2 & 675\\
REPA & SiT-XL/2 & 675\\
VA-VAE & LightningDiT-XL & 675\\
REPA-E / E2E-VAE & SiT-XL/2 + REPA & 675\\
RPiAE & LightningDiT & 675\\
FAE & Modified SiT-XL & 675\\
RAE & DiT$^{\rm DH}$-XL & 839\\
DRoRAE & DiT$^{\rm DH}$-XL & 839\\
LV-RAE & DiT$^{\rm DH}$-XL & 839\\
DecQ & DiT$^{\rm DH}$-XL & 841\\
HAE & LightningDiT-XL/1 + Riemannian FM & 677\\
\bottomrule
\end{fullwidthtabular}
\end{table}

\paragraph{Original-paper values.}
DiT/SiT benchmark rows and the SD-VAE reconstruction value are explicitly tabulated in \citet[Table 3]{yao2025vavae}; VA-VAE uses that paper's own (rFID, guided gFID) pair (0.280, 1.350). REPA uses \citet[Tables 4, 7, 9]{yu2025repa}, with interval CFG for the 1.420 guided result. REPA-E uses the E2E-VAE/REPA system from \citet[Table 9]{leng2025repae}, including its class-balanced evaluation: rFID 0.280, unguided gFID 1.690, and guided gFID 1.120. That table evaluates a generator with an already tuned E2E-VAE.

\paragraph{Image-wise reconstruction metrics.}
Table~\ref{tab:reconstruction} adds PSNR and LPIPS from each method's reported reconstruction setting. VA-VAE uses the f16d32 DINOv2-aligned row in \citet[Table 2, arXiv v3]{yao2025vavae}; REPA-E uses the VA-VAE + REPA-E row in \citet[Table 14]{leng2025repae}. RPiAE and DRoRAE use their respective main reconstruction tables, while DecQ and HAE use their Table 1 PSNR values. The SD-VAE and RAE-B PSNR/LPIPS values follow the evaluations in \citet[Table 3]{gong2026rpiae}. FAE reports neither added metric, and DecQ and HAE report no reconstruction LPIPS value. The lower block uses our matched 5K measurements with AlexNet LPIPS (Appendix~\ref{app:paired_reconstruction}).

\paragraph{Representation autoencoders.}
RAE uses the DINOv2-B/ViT-XL noise-robust tokenizer at the default $\tau=0.8$ (rFID 0.57) and DiT$^{\rm DH}$-XL results from \citet[Tables 8 and 15c]{zheng2025rae}. Its non-noise-trained decoder's 0.49 is a different configuration. RPiAE uses \citet[Table 3]{gong2026rpiae}; Section 4.2.1 describes AutoGuidance, whereas the table header says CFG. We mark this discrepancy with AG$^*$ and leave the original guided value unchanged. FAE uses the 32-channel reconstruction value in Table 9 and the timestep-shift generation rows in Table 2 of \citet{gao2025fae}. Its unguided evaluation uses 250-step SDE sampling, while guided evaluation uses 250-step ODE sampling.

\paragraph{DRoRAE baseline.}
DRoRAE~\citep{zhu2026drorae} denotes the full three-phase model, which the paper labels DRoRAE$^*$. We use rFID 0.290, guided gFID 1.650, and IS 230.600 from that model. Its tokenizer uses DINOv2-B and its DiT$^{\rm DH}$-XL generator (839M parameters) is trained for 80 epochs, with AutoGuidance scale 1.5. The corresponding unguided gFID is 2.680. The DRoRAE paper also reports RAE at 80 epochs with guided gFID 1.740 and IS 235.000. The teaser instead uses the 800-epoch RAE result, gFID 1.130 and IS 262.600; both training budgets are labeled in the main table. No six-representation aggregate was reported, so that cell remains unavailable.

\paragraph{Recent reconstruction--generation designs.}
LV-RAE uses the ``+0.1 noise'' DiT$^{\rm DH}$-XL row of \citet[Table 3]{liu2026lvrae}; its reported sampler uses 250 Euler steps and AG=1.4. Its reconstruction table reports DINO-based rFDD, so those values cannot fill an Inception rFID column. DecQ uses the default eight-query tokenizer and Tables 1--2 of \citet{wang2026decq}; its appendix specifies AutoGuidance and an Euler sampler with 50 default steps, also discussing 250 steps. HAE reconstruction and generation entries use the revised May 2026 configuration in \citet[Tables 1--3]{chang2026hae}, including its latent-smoothed tokenizer (rFID 0.78). These configuration choices matter because improving pure reconstruction can change decoder robustness during synthesis.

\paragraph{Reconstruction--generation overview sources.}
\label{app:overview}
Figure~\ref{fig:overview} (right) shows eight points: HiRAE-24, the full three-phase DRoRAE model, and six selected configurations. The RAE point uses the 800-epoch result; DRoRAE uses its 80-epoch result. Its pairs are SD-VAE/DiT (0.610, 2.270), VA-VAE/LightningDiT (0.280, 1.350), RPiAE (0.500, 1.510), RAE, 800 epochs (0.570, 1.130), DRoRAE (0.290, 1.650), DecQ (0.470, 1.050), RAEv2 (0.299, 1.060), and HiRAE-24 (0.209, 1.038). Sources and protocols are specified above and in the main tables.

\subsection{Official-checkpoint unguided evaluation}
\label{app:official_unguided}
\suppressfloats[t]
We supplement the reported RAEv2 results by evaluating the official ImageNet checkpoint from \url{https://huggingface.co/nyu-visionx/RAEv2-models}. We use 50K class-conditioned ImageNet-256 samples, 100 Euler steps, BF16, and CFG=IG=1. Sampling uses eight GPUs, 32 images per GPU, global seed 42, and rank seeds $336+\mathrm{rank}$. We compute metrics on the generated images in a separate single-GPU process.

Table~\ref{tab:unguided} uses the reported gFID 1.650 and IS 228.000 from RAEv2 Table 16~\citep{singh2026raev2}, together with $\fdr$ from our official-checkpoint evaluation.

The six normalized distances, ordered as Inception, ConvNeXt, DINOv2, MAE, SigLIP, and CLIP, are 0.997, 1.392, 2.372, 5.854, 5.698, 7.384. At the checked public release, K=23 decoder and statistics were available but its Stage-2 checkpoint was not; its missing FDr is therefore not filled with a different generator.

\subsection{Additional generation comparisons}
\label{app:additional_generation}
\suppressfloats[t]
\begin{table}[!htbp]
\centering
\caption{\textbf{ImageNet-256 generation after 80 generator epochs.} Reference systems retain their reported sampling settings; dashes denote unavailable results. $*$RPiAE's guidance label differs between the source's text and table.}
\label{tab:published80}
\small
\begin{fullwidthtabular}{4}{lrrl}
\toprule
System & Unguided gFID & Guided gFID & Guided setting\\
\midrule
RAE & 2.160 & 1.740 & AG\\
DRoRAE & 2.680 & 1.650 & AG\\
REPA-E / E2E-VAE & 3.460 & 1.670 & CFG\\
RPiAE & 2.250 & 1.510 & AG$^*$\\
FAE, timestep shift & 2.080 & 1.700 & CFG\\
DecQ, 8 queries & 1.800 & 1.330 & AG\\
HAE, revised & 2.650 & --- & ---\\
\midrule
RAEv2 & 1.650 & 1.060 & IG\\
RAEv2 K=23 & 3.010 & 1.250 & IG\\
\rowcolor{ourslight}\liftfull{} & 2.129 & 1.038 & IG\\
\bottomrule
\end{fullwidthtabular}
\end{table}

The RAE 80-epoch guided reference and DRoRAE row follow \citet{zhu2026drorae}; the other external 80-epoch rows use the cited source tables for those systems. HiRAE-24 reaches gFID 1.038 with internal guidance and 2.129 without guidance; RAEv2 reaches 1.060 and 1.650, respectively. DecQ reaches unguided gFID 1.800 after 80 generator epochs under its reported sampling protocol. Missing pixel metrics and external six-representation distances are left unfilled.

\label{app:published_fdr}
\begin{table}[!htbp]
\centering
\caption{\textbf{Guided generation measured in six representation spaces.} Baselines are transcribed from RAEv2 Table 7 and compared with HiRAE-24.}
\begin{fullwidthtabular}{4}{lrrr}
\toprule
Method & Epochs & gFID & $\fdr$\\
\midrule
SiT-XL/2 & 800 & 2.120 & 8.440\\
DDT-XL & 800 & 1.260 & 5.700\\
SiT-XL/2 + REPA & 800 & 1.420 & 5.450\\
LightningDiT & 800 & 1.420 & 4.570\\
REG & 800 & 1.540 & 4.640\\
REPA-E & 800 & 1.120 & 3.040\\
RAE-XL & 800 & 1.130 & 3.260\\
RAEv2 & 80 & 1.060 & 2.170\\
\midrule
HiRAE-24 (ours) & 80 & 1.038 & 1.856\\
\bottomrule
\end{fullwidthtabular}
\end{table}

\FloatBarrier
\section{Additional Reconstruction Results}
\label{app:additional_experiments}
\suppressfloats[t]
\subsection{Matched-image evaluation}
\label{app:paired_reconstruction}
\suppressfloats[t]
The reconstruction evaluation and mechanism diagnostics use a common cohort of 5,000 validation images from 100 fixed ImageNet classes, with 50 images per class; the feature visualizations and per-image spatial spectra use the fixed subsets specified in Appendix~\ref{app:feature_geometry}.
We keep image IDs and preprocessing fixed across tokenizers and clamp reconstructed RGB values to $[0,1]$. PSNR is computed per image from RGB mean squared error at this range; LPIPS uses the AlexNet network on images mapped to $[-1,1]$. These subset measurements complement the separate 50K rFID benchmark.

\subsection{Paired improvements and spatial-detail errors}
\label{app:reconstruction_detail}
\suppressfloats[t]
\begin{table}[!htbp]
\centering
\caption{\textbf{Paired reconstruction improvements.} HiRAE-24 minus each reference on the same 5,000 images. Intervals are 95\% within-class paired image-bootstrap intervals; negative LPIPS differences indicate improvement.}
\label{tab:paired_reconstruction_intervals}
\small
\begin{fullwidthtabular}{5}{lrrrr}
\toprule
Reference & $\Delta$PSNR & 95\% interval & $\Delta$LPIPS $\times10^3$ & 95\% interval\\
\midrule
RAEv2 & $+3.710$ & $[3.681, 3.742]$ & $-31.039$ & $[-31.327, -30.757]$\\
HiRAE-7 & $+1.309$ & $[1.292, 1.327]$ & $-6.389$ & $[-6.523, -6.257]$\\
\bottomrule
\end{fullwidthtabular}

\end{table}

We also compare spatial derivatives of the input and reconstruction. We convert images to grayscale with RGB weights $(0.299,0.587,0.114)$. Sobel error averages the squared differences in horizontal and vertical responses, using the standard $3\times3$ kernels divided by 8. Laplacian error uses the four-neighbor kernel with center weight 4 and neighbor weights $-1$. Both filters use valid convolution. Table~\ref{tab:edge_reconstruction} shows lower Sobel and Laplacian errors for HiRAE-24, consistent with the improved local detail in the selected reconstructions.

\begin{table}[!htbp]
\centering
\caption{\textbf{Reconstruction of spatial detail.} Mean derivative errors on the same matched 5K subset as Table~\ref{tab:reconstruction}, multiplied by $10^3$. Lower is better.}
\label{tab:edge_reconstruction}
\small
\begin{tabular}{lrr}
\toprule
Tokenizer & Sobel MSE $\downarrow$ & Laplacian MSE $\downarrow$\\
\midrule
RAEv2 & 1.830 & 43.378\\
HiRAE-7 & 1.167 & 36.124\\
HiRAE-24 & \textbf{0.816} & \textbf{32.645}\\
\bottomrule
\end{tabular}

\end{table}

\subsection{Reconstruction gains across texture strata}
\label{app:texture_strata}
\suppressfloats[t]
To examine which images benefit, we rank the same 5,000 inputs by their mean grayscale Sobel-gradient magnitude and split them into four quartiles of 1,250 images. Strata depend only on the original images, not on reconstruction quality or method identity. We measure both the paired LPIPS reduction from RAEv2 to HiRAE-24 and the increase caused by removing HiRAE-24's shallow residual group after its norm cap, keeping the remaining model fixed.

\begin{figure}[tb]
\centering
\includegraphics[width=\linewidth]{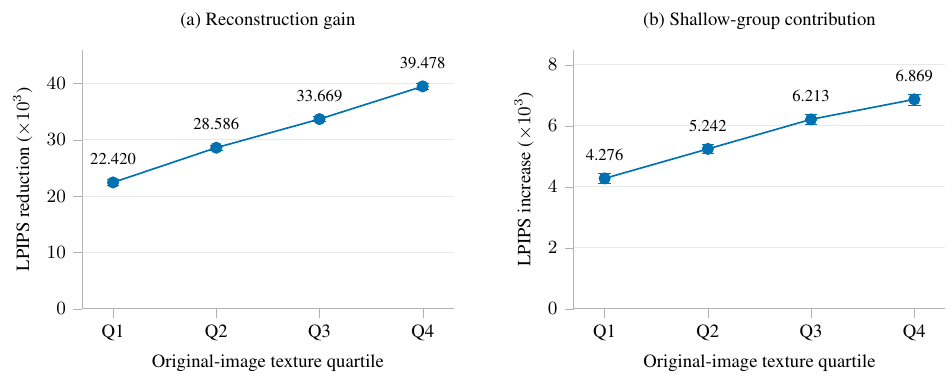}
\caption{\textbf{Texture-stratified reconstruction.} Quartiles increase in original-image Sobel strength. Left: LPIPS reduction from RAEv2 to HiRAE-24. Right: LPIPS increase after removing HiRAE-24's shallow group. Each quartile contains 1,250 images; error bars show 95\% paired bootstrap intervals.}
\label{fig:texture_strata}
\end{figure}

HiRAE-24 improves reconstruction in every quartile. From Q1 to Q4, absolute LPIPS gains rise from 0.022 to 0.039, while relative reductions decrease from 47.419\% to 37.312\%. We compute relative reduction as the mean paired LPIPS reduction divided by the mean baseline LPIPS within each quartile. Over the same quartiles, removing the shallow group increases LPIPS by 0.004 to 0.007 (Figure~\ref{fig:texture_strata}). These observations associate stronger original-image texture with a larger absolute reconstruction gain and a larger shallow-group contribution in the trained model.

All intervals use 1,000 image-resampling replicates within the fixed classes, with the same resampled IDs across paired conditions. Quartile assignments remain fixed and means are recomputed within each stratum. These intervals measure image-level sampling variability for the frozen checkpoints.

\FloatBarrier
\section{Fusion Configurations and Ablations}
\label{app:design}
\suppressfloats[t]
\subsection{HiRAE-7: selected-layer extension}
\label{app:lifae7}
\suppressfloats[t]
HiRAE-7 applies DRoRAE-style layer experts and learned aggregation within the RAEv2 framework, using the established zero-based layer subset $\{11,13,15,17,19,21,23\}$. It retains layer-specific experts, spatial routing, a deepest-layer anchor, and joint fusion--decoder training. Its router consumes concatenated normalized layer features and supplies spatially varying weights for aggregating the transformed expert outputs. Its training setup includes spatial smoothing and no router dropout. HiRAE-24 uses deepest-feature conditioning, three residual norm caps, and group dropout. The extension applies learned fusion to the selected-layer configuration.

\begin{table}[!htbp]
\centering
\caption{\textbf{Selected-layer extension of HiRAE.} Expert parameters exclude the encoder, router, decoder, and generator. rFID uses the 50K reconstruction evaluation; PSNR/LPIPS use the matched 5K subset. Generation uses 50K samples from epoch-80 generators.}
\label{tab:selected_layer_extension}
\small
\begin{tabular}{lrr}
\toprule
 & HiRAE-24 & HiRAE-7\\
\midrule
Encoder layers used & All 24 & Selected 7\\
Expert parameters (M) & $\approx$202 & $\approx$59\\
\midrule
Reconstruction rFID $\downarrow$ & 0.209 & 0.217\\
PSNR (dB) $\uparrow$ & 26.377 & 25.068\\
LPIPS $\downarrow$ & 0.043 & 0.049\\
\midrule
Guided gFID $\downarrow$ & 1.038 & 1.038\\
Guided IS $\uparrow$ & 257.823 & 257.013\\
Guided $\fdr$ $\downarrow$ & 1.856 & 1.913\\
Unguided gFID $\downarrow$ & 2.129 & 2.085\\
Unguided IS $\uparrow$ & 210.339 & 208.825\\
Unguided $\fdr$ $\downarrow$ & 4.660 & 4.479\\
\bottomrule
\end{tabular}
\end{table}

Table~\ref{tab:selected_layer_extension} shows similar rFID and guided gFID with fewer experts in HiRAE-7, while HiRAE-24 provides higher PSNR and lower LPIPS. HiRAE-24 also has lower guided $\fdr$ (1.856 versus 1.913).

HiRAE-7 achieves reconstruction rFID 0.217. On 50K guided samples, it achieves $\fdr$ 1.913, IS 257.013, and gFID 1.038. PSNR and LPIPS use the common preprocessing pipeline on the matched 5K subset.

\subsection{DRoRAE-style full-depth fusion}
\label{app:global_interpolation}
\suppressfloats[t]
The DRoRAE-style fusion comparator in Table~\ref{tab:full_depth_composition} adapts DRoRAE's layer experts, routing, and global interpolation~\citep{zhu2026drorae} to the RAEv2 framework. It uses all blocks $0,\ldots,23$ of DINOv3-L, 24 layer experts with hidden width $4C$, and the same $256\times1024$ latent interface as HiRAE-24. Its router consumes concatenated normalized layer features and produces signed, $\ell_2$-normalized weights. Writing its normalized expert combination as $F$, the output is $Z=\LN[(1-\beta)H_{23}+\beta F]$, with $\beta=0.2$. It jointly trains fusion and decoder with a routing smoothness weight of 0.1. HiRAE-24 instead uses deepest-feature routing and groupwise residual caps and dropout (Appendix~\ref{app:fusion_modules}).

Both tokenizer configurations use 16 epochs, global batch size 128, decoder-input noise $\tau=0.8$, and the ViT-XL decoder. The comparator's generator training uses global batch size 1024, GMuon, 25 warmup epochs, and a learning-rate decay endpoint at epoch 50, with the same DiT-with-DDT-head architecture. The generation results in Table~\ref{tab:full_depth_composition} use each configuration's final epoch-16 tokenizer and epoch-80 EMA generator; Table~\ref{tab:composition_budgets} additionally reports epoch-20 EMA results. Each evaluation uses 50K class-conditioned ImageNet-256 images. Their sampling configurations specify 100 Euler steps, CFG=1, and IG=1.78 on $[0.1,1]$. We use a separate generated sample set for each configuration.

This comparison evaluates complete fusion configurations with the same backbone, source layers, expert count and width, and generator architecture. HiRAE-24 achieves better guided generation with depth-dependent residual budgets during joint fusion--decoder learning. Both configurations jointly train fusion and decoding; the comparator adapts DRoRAE's fusion architecture to this setting. Its final epoch-16 tokenizer achieves reconstruction rFID 0.065498 (0.065 in Table~\ref{tab:full_depth_composition}) and also supplies the latents for generation.

Table~\ref{tab:composition_budgets} compares both fusion configurations after 20 and 80 generator epochs. At 20 epochs, the adapted DRoRAE-style fusion reaches gFID 2.244, close to HiRAE-24's 2.242, while HiRAE-24 has a 51.9\% lower $\fdr$. Extending training to 80 epochs improves this comparator to gFID 1.551 and $\fdr$ 3.375. Over the same interval, HiRAE-24 reduces gFID by 53.7\%, compared with 30.9\% for the adapted DRoRAE-style fusion, and reaches gFID 1.038 with $\fdr$ 1.856. The larger gFID improvement and lower final values of both generation metrics motivate our choice of HiRAE-24 for the main experiments.

\begin{table}[!htbp]
\centering
\caption{\textbf{Full-depth composition across training budgets.} Both configurations use all 24 layers and 24 experts. Reductions use the adapted DRoRAE-style fusion as the reference.}
\label{tab:composition_budgets}
\small
\begin{tabular*}{\linewidth}[t]{@{}l@{\extracolsep{\fill}}rrr@{}}
\toprule
Metric & \shortstack{DRoRAE-style\\fusion} & HiRAE-24 & Reduction\\
\midrule
\multicolumn{4}{@{}l}{\textit{20 generator epochs}}\\
gFID $\downarrow$ & 2.244 & \textbf{2.242} & 0.1\%\\
$\fdr$ $\downarrow$ & 5.270 & \textbf{2.533} & 51.9\%\\
\midrule
\multicolumn{4}{@{}l}{\textit{80 generator epochs}}\\
gFID $\downarrow$ & 1.551 & \textbf{1.038} & 33.1\%\\
$\fdr$ $\downarrow$ & 3.375 & \textbf{1.856} & 45.0\%\\
\bottomrule
\end{tabular*}

\end{table}

\subsection{Expert inputs and residual regularization}
\label{app:expert_variants}
\suppressfloats[t]

\paragraph{HiRAE with mode experts.}
This configuration uses all 24 inputs, partitions them into three eight-layer groups, applies a depth-axis DCT, and retains 1, 2, and 4 modes. Seven separate experts process the retained modes, with approximately 59M expert parameters compared with 202M for HiRAE-24.

Table~\ref{tab:ablations} reports three configurations at the common 20-epoch budget and the available epoch-80 results for the two regularized configurations. The Stage-1 decoder and normalization statistics are specific to each trained tokenizer.

\subsection{Number of depth groups}
\label{app:group_count}
\suppressfloats[t]
Table~\ref{tab:group_count_full} compares grouping granularity under a common depth prior. We partition the 24 DINOv3-L blocks into two, three, or four contiguous equal-sized groups. All configurations retain 24 independent layer experts with hidden width $4C$, the same router architecture and ViT-XL decoder, and the $256\times1024$ latent interface. They share initial fusion, decoder, discriminator, and generator parameters and training seed 42.

\begin{table}[!htbp]
\centering
\caption{\textbf{Three depth groups give the best generation results.} Matched budgets: 16,000 tokenizer updates and 5,004 generator updates. Reconstruction uses 5K images; each generation setting uses 10K samples.}
\label{tab:group_count_full}
\small
\begin{fullwidthtabular}{8}{rrrrrrrr}
\toprule
\multirow{2}{*}{Groups} & \multicolumn{3}{c}{Reconstruction} & \multicolumn{2}{c}{Unguided} & \multicolumn{2}{c}{IG=1.78}\\
\cmidrule(lr){2-4}\cmidrule(lr){5-6}\cmidrule(lr){7-8}
 & rFID $\downarrow$ & PSNR $\uparrow$ & LPIPS $\downarrow$ & gFID $\downarrow$ & IS $\uparrow$ & gFID $\downarrow$ & IS $\uparrow$\\
\midrule
2 & \textbf{3.293} & \textbf{22.417} & \textbf{0.1725} & 44.795 & 42.976 & 29.385 & 62.103\\
\rowcolor{ourslight}3 & 3.363 & 22.153 & 0.1754 & \textbf{44.055} & \textbf{43.216} & \textbf{28.354} & \textbf{63.796}\\
4 & 3.457 & 22.065 & 0.1789 & 45.172 & 41.618 & 31.321 & 59.984\\
\bottomrule
\end{fullwidthtabular}
\end{table}

Three groups achieve the lowest gFID and highest IS under both sampling settings, while two groups achieve slightly better reconstruction. Three groups also improve every reported metric over four groups.

\paragraph{Depth budgets and training controls.}
We map a common depth prior to each partition. Let $q_\ell$ and $p_\ell$ denote the per-layer budget and dropout probability obtained from the three-group settings: $q_\ell$ equals the original group cap divided by eight, and $p_\ell$ equals its dropout probability. For each new group $G$, we sum its layer budgets and average its dropout probabilities:
\begin{equation}
 c_G=\sum_{\ell\in G}q_\ell,\qquad
 p_G=\frac{1}{|G|}\sum_{\ell\in G}p_\ell.
\end{equation}
Table~\ref{tab:group_count_controls} lists the resulting settings in order of increasing depth. All partitions have total cap 0.25.

\begin{table}[!htbp]
\centering
\caption{\textbf{A shared depth prior across group counts.} Entries follow increasing encoder depth.}
\label{tab:group_count_controls}
\small
\begin{fullwidthtabular}{4}{rlll}
\toprule
Groups & Layers per group & Norm caps & Dropout probabilities\\
\midrule
2 & $12/12$ & $0.0625,\ 0.1875$ & $5/12,\ 0.15$\\
3 & $8/8/8$ & $0.025,\ 0.075,\ 0.15$ & $0.5,\ 0.25,\ 0.1$\\
4 & $6/6/6/6$ & $0.01875,\ 0.04375,\ 0.075,\ 0.1125$ & $0.5,\ 1/3,\ 0.2,\ 0.1$\\
\bottomrule
\end{fullwidthtabular}
\end{table}

For this ablation, the shallow, middle, and deep parts of the original prior begin opening at subset epochs 4, 2, and 0, each with a two-epoch linear ramp. If $a_\ell(e)$ is the corresponding per-layer opening coefficient, the new group uses
\begin{equation}
 a_G(e)=\frac{\sum_{\ell\in G}q_\ell a_\ell(e)}{\sum_{\ell\in G}q_\ell}.
\end{equation}
This mapping preserves the total effective cap at each training step across partitions. All groups reach their full budgets after 6,000 updates.

\paragraph{Training budget.}
Stage 1 uses a fixed, class-balanced ImageNet training subset of 128,000 images, with 128 images per class and the same sample order across configurations. Each configuration trains for 16 subset epochs at global batch size 128, totaling 16,000 updates. The optimizer, EMA, reconstruction objectives, decoder noise, and discriminator schedule follow Appendix~\ref{app:imagenet_training}, with epoch-based schedules measured on this subset. Each configuration then freezes its final EMA tokenizer and computes its own latent normalization statistics on the same 128,000 images.

Stage 2 trains each generator from the shared initialization on all 1,281,167 ImageNet training images for four epochs, totaling 5,004 updates. Each run uses eight H800 GPUs, microbatch size 64 per GPU, two accumulation steps, and global batch size 1024. GMuon uses a constant learning rate of $2\times10^{-4}$ from the first update, momentum 0.95, Nesterov acceleration, and zero weight decay; gradient clipping is 1.0 and EMA decay is 0.9995. We retain the internal-guidance base-branch loss and use each stage's final EMA weights for evaluation.

\paragraph{Evaluation.}
Reconstruction uses a fixed class-balanced ImageNet validation subset with five images per class, totaling 5,000 images. The rFID reference contains these same original images; PSNR and LPIPS use paired originals and reconstructions. All residual groups operate at their full budgets, with dropout and decoder-input noise disabled.

For generation, each setting produces 10,000 images with ten per class, using 50 Euler steps. All configurations share the class sequence, sampling seed 42, eight GPUs, and batch size 16 per GPU. Unguided sampling uses CFG=IG=1; guided sampling uses CFG=1 and IG=1.78 on $[0.1,1.0]$. We compute gFID and IS with the same project Inception implementation and ImageNet reference statistics from \texttt{guided\_diffusion\_stats.npz}.

\subsection{Layer-group and frequency interventions}
\label{app:group_interventions}
\suppressfloats[t]
We use the matched-image cohort in Appendix~\ref{app:paired_reconstruction}. We modify each group residual after its norm cap and before the final LayerNorm:
\begin{equation}
 z(\boldsymbol{\alpha})=\LN\!\left(H_{23}+\alpha_s\Delta_s+\alpha_m\Delta_m+\alpha_d\Delta_d\right).
\end{equation}
Setting one coefficient to zero removes that group's contribution. We retain the other groups, routing weights, and decoder, with no redistribution or fine-tuning. Table~\ref{tab:group_removal} reports the increase in reconstruction LPIPS relative to the unmodified tokenizer.

\begin{table}[!htbp]
\centering
\caption{\textbf{Each residual group contributes to reconstruction.} We remove one group from the frozen HiRAE-24 tokenizer. Values report $\Delta\mathrm{LPIPS}\times10^3$ with paired 95\% intervals.}
\label{tab:group_removal}
\begin{tabular}{lrrr}
\toprule
Group & Norm cap & LPIPS increase & Paired interval\\
\midrule
Shallow (0--7) & 0.025 & 5.650 & [5.563, 5.737]\\
Middle (8--15) & 0.075 & 54.111 & [53.600, 54.626]\\
Deep (16--23) & 0.150 & 147.750 & [146.789, 148.661]\\
\bottomrule
\end{tabular}
\end{table}

Frequency interventions use an orthonormal spatial DCT on the $16\times16$ token grid. For DCT indices $p,q$, the low, middle, and high bands satisfy $0<p+q\leq10$, $10<p+q\leq20$, and $p+q>20$, respectively. We retain the zero-frequency (DC) coefficient. One control compares high-frequency removal with uniform shrinkage that leaves the same total non-DC energy in that group. A second removes equal-norm low- or high-frequency components, with perturbation norm $q_g=\kappa\min(\|\Delta_{g,\mathrm{low}}\|_F,\|\Delta_{g,\mathrm{high}}\|_F)$ for each image and group $g$.

\begin{table}[!htbp]
\centering
\caption{\textbf{Depth groups differ in frequency contributions.} Values report differences in reconstruction LPIPS, multiplied by $10^3$. The equal-norm comparison uses $\kappa=1$; positive values indicate greater damage from high-frequency removal.}
\label{tab:frequency_interventions}
\small
\begin{tabular}{lrrr}
\toprule
Group & High removal $-$ energy control & High $-$ low, equal norm & Paired interval\\
\midrule
Shallow & 0.770 & 0.215 & [0.182, 0.248]\\
Middle & 4.760 & 0.068 & [$-0.013$, 0.144]\\
Deep & 7.328 & $-2.053$ & [$-2.163$, $-1.941$]\\
\bottomrule
\end{tabular}
\end{table}

Table~\ref{tab:frequency_interventions} shows a greater effect from high-frequency removal in the shallow group and from low-frequency removal in the deep group. The middle-group interval includes zero. At $\kappa=0.5$, the shallow and deep groups retain their respective directions, while the middle group weakly favors high-frequency removal (0.109, interval [0.070, 0.149]). We obtain paired intervals from 1,000 image-resampling replicates within the fixed classes, sharing image IDs across conditions.

\FloatBarrier
\section{Latent-Space and Decoding Analysis}
\label{app:mechanism}
\subsection{Analysis protocol}
\label{app:analysis_protocol}
We use the matched-image cohort in Appendix~\ref{app:paired_reconstruction}; the feature visualizations and spatial spectra use the subsets specified in Appendix~\ref{app:feature_geometry}. We freeze the tokenizers and generators; the HiRAE-24 generator uses its epoch-80 EMA checkpoint. Decoding and generator diagnostics use each system's own latent-normalization statistics; the feature-geometry analysis specifies its separate coordinate conventions. Within-model interventions keep the remaining modules fixed. The experiments therefore require no additional training.

\subsection{Class organization and spatial enrichment}
\label{app:feature_geometry}
\paragraph{Class organization and spatial effective rank.}
Figure~\ref{fig:latent_analysis}(a) uses raw tokenizer coordinates: spatially pooled image vectors are $\ell_2$-normalized without additional channel standardization. We select ten classes at evenly spaced positions in the sorted 100-class list and retain all 50 images per selected class. PHATE~\citep{moon2019phate} is fitted separately for each model with 10 neighbors, decay 40, an internal 50-dimensional PCA, automatically selected diffusion time, metric MDS, and random seed 42. We interpret class neighborhoods separately in each PHATE embedding. The accompanying cosine 10NN statistic uses all 5,000 images, excluding each query's own ID, and is computed before dimensionality reduction. Under these coordinates, the HiRAE-24 anchor has a same-class fraction of 79.634\%, compared with 79.536\% after fusion. Applying shared anchor-channel standardization changes RAEv2's fraction from 74.454\% to 66.024\%, and HiRAE-24's from 79.536\% to 81.466\%.

Figure~\ref{fig:latent_analysis}(b) instead compares the deep anchor and fusion \emph{within HiRAE-24}. We take the image at zero-based within-class position 25 from each class, giving 100 fixed images. For each $256\times1024$ feature matrix, we center channels over spatial tokens and compute singular values $\sigma_j$. Spatial effective rank is $\exp(-\sum_j p_j\log p_j)$, where $p_j=\sigma_j/\sum_k\sigma_k$; its maximum is 255 after spatial centering. Spatial CKA~\citep{kornblith2019similarity} is centered linear CKA between the paired token Gram matrices, averaged across images. The mean paired rank increase is 24.533, with 95\% CI [23.832, 25.264]; mean spatial CKA is 0.985, with CI [0.983, 0.986]. We compute intervals from 2,000 bootstrap resamples of the 100 whole image pairs, with one fixed image per class. Effective rank summarizes the distribution of spatial variation across feature directions; the reconstruction interventions assess the contributions of the depth groups.

\paragraph{Anchor-to-fusion representation structure.}
We compare $\LN(H_{23})$ with the final fused representation using one set of channel means and standard deviations fitted to the normalized anchor. We also report results in raw coordinates. Spatial high-frequency energy uses DCT indices $p+q>15$, with the DC coefficient excluded from the denominator. Class-neighborhood consistency measures the mean same-class fraction among ten nearest neighbors of globally pooled image representations, using a fixed gallery and excluding each query's own ID. Pooling precedes image-wise spatial centering.

\begin{table}[!htbp]
\centering
\caption{\textbf{Spatial enrichment largely preserves class neighborhoods.} Values compare HiRAE-24's deep anchor and fused representation under shared anchor normalization and without additional standardization.}
\label{tab:anchor_structure}
\small
\begin{tabular}{llrr}
\toprule
Coordinates & Representation & High-frequency fraction & Same-class 10NN\\
\midrule
Shared anchor & Deep anchor & 0.123 & 0.827\\
Shared anchor & Fused representation & 0.208 & 0.815\\
\midrule
Raw & Deep anchor & 0.107 & 0.796\\
Raw & Fused representation & 0.166 & 0.795\\
\bottomrule
\end{tabular}
\end{table}

Both coordinate choices in Table~\ref{tab:anchor_structure} show increased spatial high-frequency content and a smaller change in class neighborhoods. These neighborhood statistics use the fixed 100-class gallery.

\subsection{Cross-system spatial statistics}
\label{app:spatial_comparison}
\begin{figure}[tb]
\centering
\includegraphics[width=0.60\linewidth]{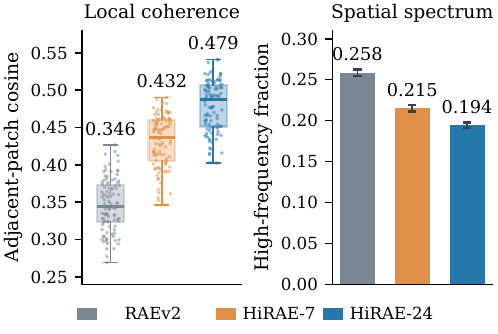}
\caption{\textbf{Cross-system spatial statistics.} Left: class-mean adjacent-patch cosine similarity. Right: high-frequency DCT energy fraction without DC. We standardize each representation per channel and remove spatial means for the cosine measure.}
\label{fig:latent_spatial_comparison}
\end{figure}

The cross-system analysis follows latent-diffusability diagnostics~\citep{zhong2026diffusability}, using each representation's own channel standardization. After image-wise spatial centering, adjacent-patch cosine similarity is 0.346, 0.432, and 0.479 for RAEv2, HiRAE-7, and HiRAE-24; high-frequency energy fractions are 0.258, 0.215, and 0.194. Figure~\ref{fig:latent_spatial_comparison} summarizes these spatial measurements. The cross-system statistics use each tokenizer's standardized coordinates; the anchor-to-fusion statistics use shared anchor normalization within HiRAE-24.

\subsection{Decoding response to latent perturbations}
\label{app:decoding_response}
Let $u$ denote the standardized latent that the generator models, and let $G$ include inverse normalization and the decoder. We measure
\begin{equation}
 S(u,\delta)=\operatorname{LPIPS}\!\left(G(u),G(u+\delta)\right),
 \qquad \|\delta\|_F/\|u\|_F\in\{0,0.01,0.02,0.05,0.10\}.
\end{equation}
We use Gaussian, spatial low-frequency, and spatial high-frequency directions, with three fixed seeds per direction type. We average directions within each image and clip decoded outputs to $[0,1]$ before computing LPIPS. Figure~\ref{fig:latent_analysis}(c) shows the 10\% endpoint; Table~\ref{tab:decoder_response} provides its numerical values.

\begin{table}[!htbp]
\centering
\caption{\textbf{HiRAE decodings change less under the tested perturbations.} Output LPIPS $\times10^3$ at 10\% relative latent perturbation norm.}
\label{tab:decoder_response}
\begin{tabular}{lrrr}
\toprule
Tokenizer & Gaussian & Low frequency & High frequency\\
\midrule
RAEv2 & 8.385 & 7.191 & 9.414\\
HiRAE-7 & 1.576 & 1.344 & 1.810\\
HiRAE-24 & 1.791 & 1.547 & 2.043\\
\bottomrule
\end{tabular}
\end{table}

Both HiRAE configurations produce smaller decoded changes than RAEv2 under all three perturbation types, with HiRAE-7 showing the smaller response.

\subsection{Generator prediction error}
\label{app:generator_error}
We encode matched real images with each tokenizer and add noise at seven common coefficient log-SNR levels, fixing noise by image ID. We measure the full prediction branch's clean-latent mean squared error (MSE) per coefficient, including DC. This diagnostic measures prediction error on forward-noised real-image latents.

\begin{figure}[tb]
\centering
\includegraphics[width=0.65\linewidth]{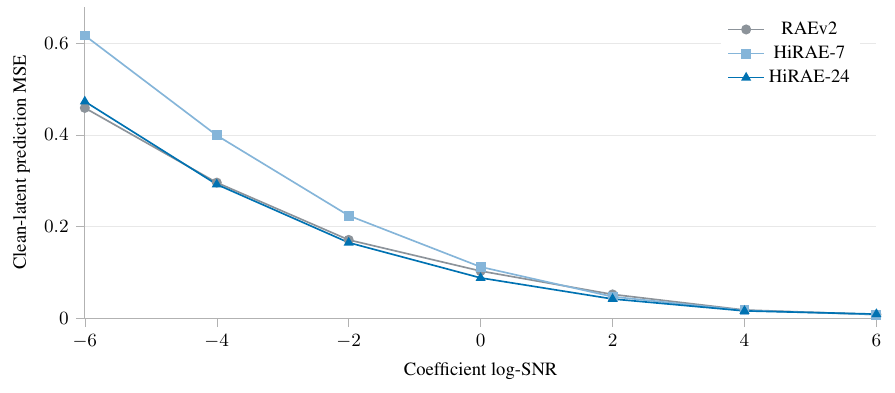}
\caption{\textbf{Prediction error varies with noise level.} Full-branch clean-latent MSE on matched forward-noised images. HiRAE-24 improves over RAEv2 at the five interior levels; both endpoints have slightly higher error.}
\label{fig:denoising_diagnostics}
\end{figure}

\begin{table}[!htbp]
\centering
\caption{\textbf{Complete prediction-error grid.} Per-coefficient clean-latent MSE, including DC, at all seven coefficient log-SNR levels.}
\label{tab:denoising_diagnostics}
\begin{tabular}{rrrr}
\toprule
Coefficient log-SNR & RAEv2 & HiRAE-7 & HiRAE-24\\
\midrule
$-6$ & 0.459 & 0.617 & 0.473\\
$-4$ & 0.296 & 0.399 & 0.292\\
$-2$ & 0.171 & 0.224 & 0.165\\
0 & 0.103 & 0.112 & 0.088\\
2 & 0.052 & 0.047 & 0.042\\
4 & 0.018 & 0.017 & 0.016\\
6 & 0.008 & 0.008 & 0.009\\
\bottomrule
\end{tabular}
\end{table}

The five-interior-level ordering persists after dividing each system's prediction MSE by its own target energy. At log-SNR 0, HiRAE-24 also has lower per-coefficient error in each of the low, middle, and high frequency bands. The log-SNR 6 endpoint corresponds to $t\approx0.047$; the final Euler100 model evaluation occurs at $t\approx0.075$.

\FloatBarrier
\section{Additional Qualitative Results}
\label{app:qualitative}
\suppressfloats[t]
\subsection{Reconstruction comparisons}
\label{app:qualitative_reconstruction}
\begin{figure}[H]
\centering
\includegraphics[width=0.92\linewidth,height=0.64\textheight,keepaspectratio]{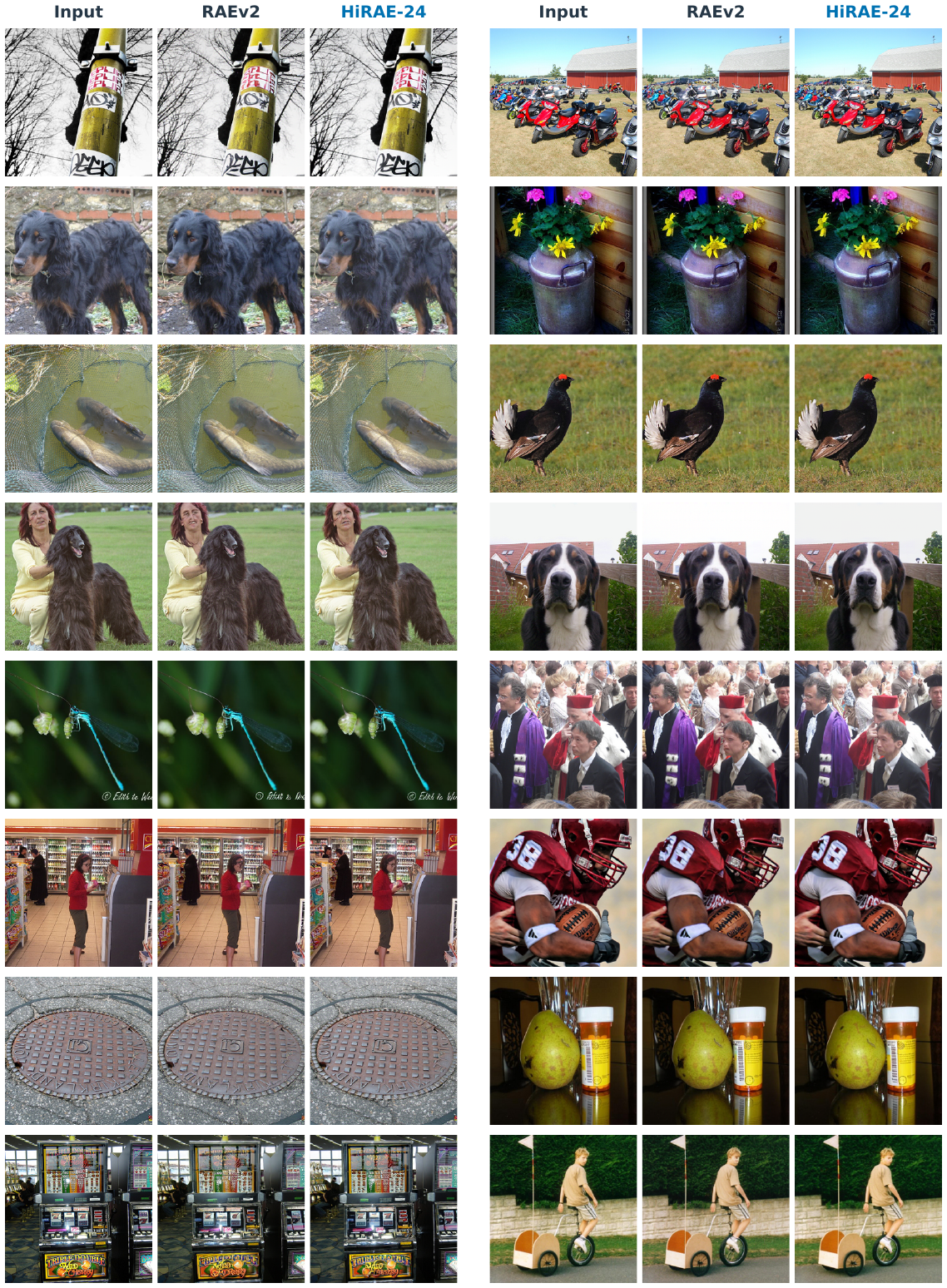}
\caption{\textbf{Additional reconstruction comparisons.} Each triplet shows the input, official RAEv2 reconstruction, and HiRAE-24 reconstruction. The first two rows contain four selected cases; the remaining six rows contain twelve cases at predetermined validation indices.}
\label{fig:selected_appendix}
\end{figure}

\suppressfloats[t]
Figure~\ref{fig:selected_appendix} compares additional reconstructions from RAEv2 and HiRAE-24 on selected and predetermined inputs.

\clearpage
\subsection{Generation samples}
\label{app:qualitative_generation}
\begin{figure}[!htbp]
\centering
\includegraphics[width=0.86\linewidth]{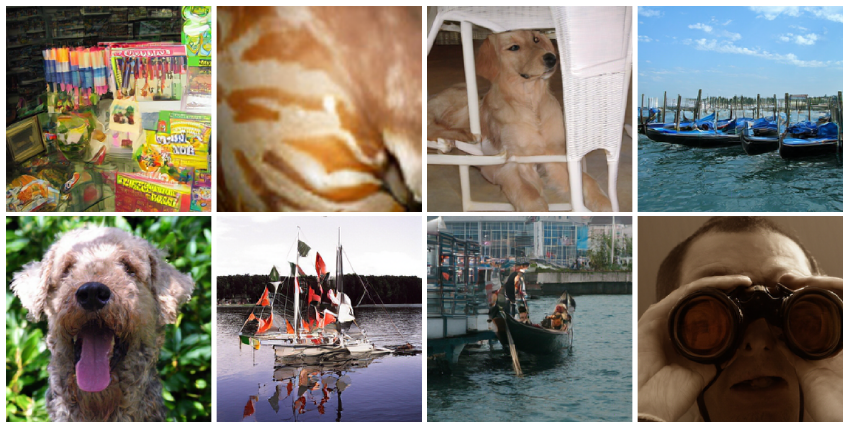}
\caption{\textbf{Unguided class-conditioned samples from HiRAE-24.} The first eight stored samples are shown without quality filtering, using the epoch-80 EMA model with CFG=IG=1.}
\label{fig:unguided_appendix}
\end{figure}

\suppressfloats[t]
Figure~\ref{fig:unguided_appendix} shows the first eight stored unguided samples from HiRAE-24. Figure~\ref{fig:guided_samples} in the main paper shows selected guided samples.

\FloatBarrier
\subsection{Sample selection and visualization details}
\label{app:sample_selection}
\suppressfloats[t]
\paragraph{Reconstruction examples.}
Reconstruction inference uses the epoch-16 EMA HiRAE-24 checkpoint and the official ImageNet RAEv2 decoder with its corresponding frozen DINOv3-L aggregation encoder. We use BF16, batches of four, and no decoder-input noise. Outputs are clamped to $[0,1]$ and converted to uint8 identically for both models. We select reconstruction examples through pixel-error screening and visual inspection of ImageNet validation images.

Figure~\ref{fig:overview} uses index 21673. Figure~\ref{fig:reconstruction} uses indices 47622 and 33692 in left-to-right triplet order. The two main examples show complete images with shared red boxes and enlarged crops below. In the same order, the crop coordinates $(x_0,y_0,x_1,y_1)$ are $(0,10,88,98)$ and $(102,16,193,89)$. Coordinates refer to the original $256\times256$ images, with a top-left origin and exclusive right/bottom boundaries. Each triplet uses identical coordinates and nearest-neighbor enlargement, preserving the crop aspect ratio. The teaser uses the shared crop $(65,14,142,91)$, a $77\times77$ region. Its input context image is uniformly scaled and horizontally cropped to a portrait viewport; the three enlarged detail crops retain their shared square region. The selected reconstruction examples illustrate local detail; the matched 5K reconstruction metrics quantify average image-wise performance. The four additional selected examples, at indices 36681, 33541, 10731, and 32659, are retained in Figure~\ref{fig:selected_appendix} as complete images, without red boxes or enlarged crops. The same figure also includes the predetermined 12-image set at indices $0,4000,\ldots,44000$.

\paragraph{Generation examples.}
Figure~\ref{fig:guided_samples} shows 24 examples selected by visual inspection from the first 384 entries of the guided 50K archive. Eight groups organize the examples by visible subject matter, each placing a representative beside two related examples. We display complete images without cropping, sharpening, or color adjustment. The unguided grid shows the first eight entries of its separate archive.

\clearpage
\subsection{Text-to-image comparisons}
\label{app:qualitative_t2i}
\begin{figure}[H]
\centering
\includegraphics[width=0.94\linewidth,height=0.80\textheight,keepaspectratio]{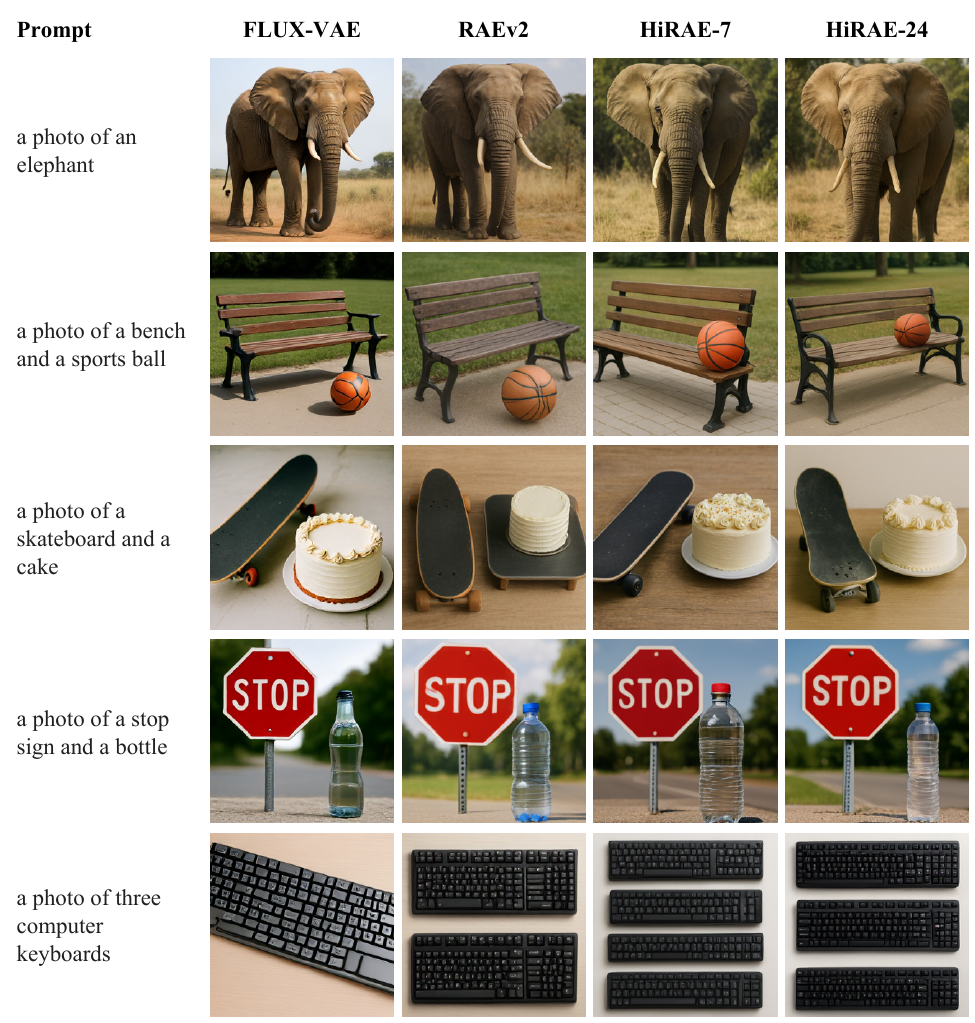}
\caption{\textbf{Qualitative comparisons on five selected GenEval prompts.} Each row shows the prompt and the outputs of FLUX-VAE, RAEv2, HiRAE-7, and HiRAE-24 after SFT. We display the complete original images. FLUX-VAE denotes the FLUX autoencoder paired with our trained DiT.}
\label{fig:geneval_selected}
\end{figure}

\end{document}